\documentclass[10pt]{article} % For LaTeX2e
\usepackage[preprint]{tmlr}
\usepackage{amsmath,amsfonts,amssymb,bm}
\usepackage{hyperref}
\usepackage{url}
\usepackage{graphicx}
\usepackage{booktabs}
\usepackage{multirow}
\usepackage{wrapfig}
\usepackage{subcaption}
\usepackage{xcolor}

\hypersetup{colorlinks=true, linkcolor=blue, citecolor=blue, urlcolor=blue}

\renewcommand{\headrulewidth}{0.4pt}
\fancypagestyle{firstpage}{%
  \fancyhf{}%
  \fancyfoot[L]{\textit{Technical Report}}%
  \fancyfoot[C]{\thepage}%
  \renewcommand{\headrulewidth}{0.4pt}%
}

\newcommand{\af}{\alpha_{\mathrm{f}}}
\newcommand{\ac}{\alpha_{\mathrm{c}}}
\newcommand{\atk}{\alpha_{\mathrm{tK}}}
\newcommand{\LCE}{\mathcal{L}_{\mathrm{CE}}}
\newcommand{\Lfe}{\mathcal{L}_{\mathrm{fe}}}
\newcommand{\Rchain}{R_{\mathrm{chain}}}
\newcommand{\RtopK}{R_{\mathrm{topK}}}
\newcommand{\taubar}{\bar{\tau}}
\newcommand{\E}{\mathbb{E}}
\newcommand{\Hlt}{H_{<t}}
\newcommand{\None}{\texttt{None}}

\newcommand{\papertitle}{Teaching Diffusion to Speculate Left-to-Right}
\title{\papertitle}

\author{\name Lexington Whalen\thanks{Corresponding author.} \email lexington.whalen@sbintuitions.co.jp \\
      \addr SB Intuitions
      \AND
      \name Yuki Ito \email yuki.ito@sbintuitions.co.jp \\
      \addr SB Intuitions
      \AND
      \name Ryo Sakamoto \email ryo.sakamoto@sbintuitions.co.jp \\
      \addr SB Intuitions}

\def\month{MM}  % Insert correct month for camera-ready version
\def\year{YYYY} % Insert correct year for camera-ready version
\def\openreview{\url{https://openreview.net/forum?id=XXXX}} % Insert correct link to OpenReview for camera-ready version

\begin{document}

\maketitle
\thispagestyle{firstpage}

\begin{abstract}
Speculative decoding accelerates large language model inference by having a lightweight drafter propose blocks of candidate tokens that a larger target verifies in a parallel pass under a rejection-sampling contract that preserves the target's output distribution exactly. Block-diffusion drafters are attractive in this setting because they emit an entire $B$-token block in one non-autoregressive pass, but they are trained with fully bidirectional attention within the block, whereas verification proceeds strictly left-to-right and truncates the block at the first rejection. We show that this training--verification mismatch is empirically consequential: on a position-uniform block-diffusion baseline, an average of 46.9\% of draft tokens that match the target's greedy output are discarded as a downstream consequence of an earlier within-block rejection. We then analyse three complementary training-time interventions that reshape the drafter's objective along orthogonal axes: position-wise loss decay, a first-error focal term targeting the block's chain-breaking position, and a chain reward that substitutes a differentiable surrogate for the expected accepted length. We show that they compose additively, add negligible compute, and require no change to the drafter architecture, the inference pipeline, or the exactness contract. Across four instruction-tuned targets (Llama-3.2-3B, Llama-3-8B, Qwen3-4B, Qwen3-8B) and six reasoning, code, and dialogue benchmarks, the fully stacked configuration raises average accepted draft length by up to +43.9\% over the position-uniform baseline and by up to +75.9\% per individual benchmark. The interventions further compound when composed with target-aligned training data, varied speculation horizons, tree-based verification, and the pathwise streak-distillation objective of SpecDiff-2, indicating that they act on an axis distinct from data, horizon, and verifier-side alignment.
\end{abstract}

% ===========================================================================
\section{Introduction}
% ===========================================================================

Large language models (LLMs) have become a central computational primitive across applications ranging from conversational assistants and code generation \citet{chen2021codex,roziere2023codellama} to retrieval-augmented question answering, multi-step agent systems, and long-horizon reasoning \citet{openai2024o1,deepseek2025r1}. As these systems mature, the dominant cost in their lifecycle has shifted from training to inference: a model is trained once but served continuously, and per-query cost is amplified by output length, by recursive agentic invocations, and by the long chains of thought emitted by recent reasoning models. Reducing the latency and per-token cost of LLM inference is therefore one of the most consequential systems problems in contemporary machine learning.

Single-stream LLM decoding is bottlenecked by memory bandwidth rather than arithmetic throughput: each autoregressive step streams the entire parameter set from high-bandwidth memory (HBM) to produce a single token, an imbalance that widens with every hardware generation as compute scales faster than HBM \citet{pope2023efficiently}. The standard responses---quantization \citet{dettmers2022llmint8,frantar2023gptq,lin2024awq}, sparsity \citet{frantar2023sparsegpt}, distillation \citet{hinton2015distilling}, and system-level techniques such as FlashAttention \citet{dao2022flashattention}, PagedAttention, and continuous batching \citet{kwon2023vllm,yu2022orca}---all reduce the cost of a single forward pass. Speculative decoding \citet{leviathan2023fast,chen2023sps} instead reduces the number of serialized forward passes required: a cheap drafter proposes $B$ candidate tokens, the target verifies them in a single parallel pass, and a rejection-sampling step accepts the longest prefix consistent with the target distribution. The procedure is ensures samples are statistically identical to those of the target model, which eliminates the quality--versus--speed trade-off that complicates the other techniques and makes the central question purely one of drafter design: how to construct a draft distribution that is simultaneously cheap to evaluate and well aligned with the target.

\begin{figure}[t]
  \centering
  \includegraphics[width=0.75\linewidth]{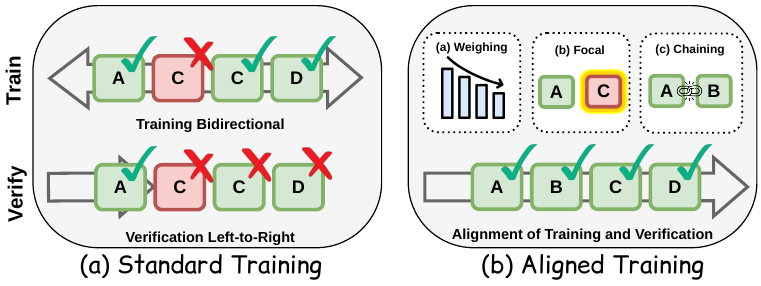}
  \caption{The training--verification mismatch in diffusion-based speculative drafters and the three training-time interventions we analyse. \textbf{(a) Standard training.} The drafter is trained with fully bidirectional attention over the block, so every position is conditioned symmetrically on every other. At inference time, however, the target verifies the block strictly left-to-right: a single early rejection (here, position 2) truncates the entire suffix. The drafter's \texttt{accd} matches the ground-truth \texttt{abcd} at positions 3 and 4, but both are discarded along with the rejected position-2 token. \textbf{(b) Aligned training.} We analyze three complementary training-time interventions that bring the drafter's objective into closer correspondence with the causal acceptance contract: position-wise loss decay (\emph{Weighting}), a term targeting the first mispredicted position of each block (\emph{Focal}), and a differentiable surrogate for the joint prefix acceptance probability (\emph{Chaining}). Together they raise the expected accepted prefix length and recover the per-token correctness that bidirectional training already provides.}
  \label{fig:overview}
\end{figure}

The EAGLE family \citet{li2024eagle,li2024eagle2,li2025eagle3} pursues this goal by drafting in \emph{feature} space rather than token space, coupling a small auxiliary head to the target's own hidden states. Successive iterations have introduced dynamic tree expansion and multi-layer feature aggregation, and EAGLE-3 has become the de facto baseline for production speculative decoding. The EAGLE drafter nevertheless remains \emph{autoregressive} within each speculation step, requiring token predictions to be generated sequentially and thereby limiting its maximum achievable speedup.

In this work we focus on \emph{block-diffusion drafters}, of which DFlash \citep{chen2026dflash} is a prominent recent example. This class of speculative decoding methods replaces autoregressive feature prediction with a \emph{diffusion-style} parallel block decoder. Conditioned on a intermediate target hidden states, such a drafter emits an entire $B$-token block in a single forward pass. Pairing a block-diffusion drafter with an autoregressive target also surfaces a representational asymmetry: diffusion models are trained to denoise blocks under fully bidirectional attention, so each position conditions on context from both directions, whereas the target verifies strictly left-to-right and accepts only the longest causally consistent prefix. The drafter should therefor allocate its predictive capacity assymetrically, since a single early divergence truncates all subsequent draft tokens regardless of their quality---an objective in tension with the symmetric denoising loss inherited from the diffusion formulation. The present work studies this tension by analyzing three training-time interventions that reshape the position-wise loss profile along different axes; the analysis is complementary to test-time alignment mechanisms for the drafter class, such as ddTree \citet{ringel2026ddtree}.

\paragraph{Contributions.}
\begin{itemize}
  \item We isolate and quantify a specific failure mode of block-diffusion drafters under standard verification: on a uniformly-weighted DFlash baseline, an average of 46.9\% of draft tokens that match the target's greedy output are nevertheless discarded by the rejection-sampling contract as a consequence of upstream within-block rejections (Section~\ref{sec:gap}).
  \item We formalise three complementary training-time interventions that reshape the drafter's loss along orthogonal axes---position (loss decay), block-conditional first error (first-error focal loss), and joint prefix (chain reward)---and show that they compose additively without altering the drafter architecture, the inference pipeline, or the rejection-sampling exactness contract (Section~\ref{sec:techniques}).
  \item Across four instruction-tuned targets (Llama-3.2-3B, Llama-3-8B, Qwen3-4B, Qwen3-8B) and six reasoning, code, and dialogue benchmarks, the fully stacked configuration raises average accepted length by +20.9 to +43.9\% over the position-uniform baseline, with per-benchmark gains as large as +75.9\% (Section~\ref{sec:other-models}).
  \item We show how the three interventions work with axes the training loss does not directly control: target-aligned training corpora (Section~\ref{sec:data}), inference-time speculation horizon (Section~\ref{sec:blocksize}), tree verification (Section~\ref{sec:ddtree}), and the pathwise streak-distillation objective of SpecDiff-2 (Section~\ref{sec:specdiff}). In every case, the ranking and compounding pattern of the training-time interventions is preserved, showing that they operate on a distinct axis from data, horizon, and verifier-side alignment.
\end{itemize}

% ===========================================================================
\section{Related Work}
% ===========================================================================

\paragraph{Efficient LLM inference.} A broad portfolio of techniques mitigates the memory-bandwidth cost of single-stream decoding: low-bit quantization \citet{dettmers2022llmint8,frantar2023gptq,lin2024awq,xiao2023smoothquant}, sparsity \citet{frantar2023sparsegpt,sun2024wanda}, distillation \citet{hinton2015distilling,sanh2019distilbert}, FlashAttention \citet{dao2022flashattention,dao2024flashattention2}, and system-level techniques such as PagedAttention, continuous batching, and chunked prefill \citet{kwon2023vllm,yu2022orca,agrawal2023sarathi}.

\paragraph{Speculative decoding.} Block-parallel decoding was first explored as a deterministic acceleration technique \citet{stern2018blockwise}; the modern, distribution-preserving formulation was introduced concurrently by \citet{leviathan2023fast} and \citet{chen2023sps}. SpecInfer \citet{miao2024specinfer} generalized linear-chain drafts to trees verified in parallel under tree-structured attention, substantially raising expected accepted lengths. Medusa \citet{cai2024medusa} attaches parallel prediction heads directly to the target, Lookahead Decoding \citet{fu2024lookahead} sidesteps drafter training via Jacobi-style fixed-point iteration on $n$-gram trajectories, self-speculative decoding \citet{zhang2024draftverify} reuses a subset of target layers as the drafter, and online speculative decoding \citet{liu2024online} adapts the drafter continuously to the deployed workload.

\paragraph{Feature-level drafters.} A particularly successful line draws candidates from the target's own intermediate representations rather than from an independently trained small LM. EAGLE \citet{li2024eagle} introduced feature-level autoregression; EAGLE-2 \citet{li2024eagle2} added context-dependent dynamic tree expansion; and EAGLE-3 \citet{li2025eagle3} augmented the drafter with multi-layer feature aggregation, establishing the de facto baseline for production speculative decoding. Related designs include Hydra \citet{ankner2024hydra}, GliDe with CaPE \citet{du2024glide}, Kangaroo \citet{liu2024kangaroo}, and HASS \citet{zhang2024hass}. All remain autoregressive within a speculation step, so their empirical horizons are bounded by tree depth rather than by the cost of a single drafter pass.

\paragraph{Diffusion and non-autoregressive language modelling.} Parallel-block generation has a long history outside speculative decoding: non-autoregressive translation \citet{gu2018nat} first demonstrated single-pass sequence generation, and iterative refinement \citet{lee2018iterative} together with masked-LM decoders such as Mask-Predict \citet{ghazvininejad2019maskpredict} and SUNDAE \citet{savinov2022sundae} recovered much of the resulting quality gap via repeated denoising. Discrete diffusion LMs \citet{austin2021d3pm,li2022diffusionlm,gong2023diffuseq,gulrajani2023plaid,lou2024sedd,nie2025llada} scaled this paradigm to LLM size via bidirectional denoising of masked blocks, and block diffusion \citet{arriola2025block} interpolates between this regime and autoregression by combining within-block bidirectional denoising with between-block causal conditioning. More recent work targets diffusion LMs as efficient generators in their own right: EfficientDLM \citet{fu2026efficientdlm} uses position-dependent token masking; TiDAR \citet{liu2025tidar} casts a single model as both drafter and verifier; and Nemotron-Labs-Diffusion \citet{fu2026nemotron} introduces a tri-modal architecture unifying autoregressive, diffusion, and self-speculative decoding. The natural synthesis pursued in this work---block-diffusion models as speculative drafters, of which DFlash \citep{chen2026dflash} is a prominent recent example---inherits the parallel-block speedup of these models while introducing, as we show in Section~\ref{sec:gap}, a training objective in tension with the causal acceptance contract.

\paragraph{Aligning drafters with the acceptance contract.} A rapidly growing literature targets the same gap we do; the present work is distinguished by (i) its focus on the bidirectional-vs-causal mismatch specific to block-diffusion drafters (rather than autoregressive Medusa/EAGLE heads), (ii) three composable interventions covering three distinct axes of the loss surface (position, block-conditional first error, joint prefix) rather than a single loss family, and (iii) empirical composition with, rather than substitution for, contemporaneous train-time (SpecDiff-2, D-PACE) and test-time (DDTree) alignment mechanisms. \citet{zhou2024distillspec} study reverse-KL and total-variation alternatives to the standard forward-KL distillation loss for autoregressive drafters, and the recent LK losses of \citet{samarin2026lk} optimize a TV-based per-token acceptance objective. These methods change the per-token loss family but weight all $B$ draft positions uniformly and remain agnostic to the joint-prefix structure of the acceptance contract; they are also developed against autoregressive drafters (Medusa-style heads, EAGLE variants, MTP modules) rather than bidirectional block decoders. SpecDiff-2 of \citet{sandler2025specdiff2}, targets the same block-diffusion-drafter / autoregressive-verifier mismatch via streak-distillation---a fine-tuning objective maximising a differentiable surrogate for the expected accepted streak under verifier-sampled teacher trajectories---paired with a test-time self-selection acceptance mechanism. The chain reward (Section~\ref{sec:chain}) shares streak-distillation's animating idea but evaluates the surrogate along the teacher-forced ground-truth tokens already materialized by $\LCE$, adding only a cumulative sum and exponentiation per block---no verifier rollout (Section~\ref{sec:specdiff} shows that we can use both). A contemporaneous D-PACE \citep{wu2026dpace} also targets this gap with a prefix-product accepted-length surrogate similar to the chain reward we investigate, but applies its gradient as a detached per-position weight on the standard cross-entropy rather than as an additive reward term. The present analysis is otherwise scoped entirely to training-time interventions and leaves the inference pipeline at the standard rejection-sampling contract; test-time mechanisms such as the tree-based selection of DDTree \citep{ringel2026ddtree} are orthogonal and can be combined with the interventions we explore ( Sections~\ref{sec:ddtree}--\ref{sec:specdiff}).

% ===========================================================================
\section{Preliminaries}
\label{sec:prelim}
% ===========================================================================

\subsection{Notation}

We consider an autoregressive target model $p_\phi(x_t \mid x_{<t})$ over a vocabulary $\mathcal{V}$ with parameters $\phi$, generating sequences $x_{1:T}$ according to $p_\phi(x_{1:T}) = \prod_{t=1}^{T} p_\phi(x_t \mid x_{<t})$. We write $h^{(\ell)}_{1:t}$ for the hidden states produced by the $\ell$-th transformer block of $p_\phi$ when conditioned on $x_{<t}$. A drafter is a parametric distribution $q_\psi(x_t \mid x_{<t})$ designed to approximate $p_\phi$ at lower per-token cost.

\subsection{Speculative Decoding}

Given a prefix $x_{<t}$ and a draft horizon $B \in \mathbb{N}$, one iteration of speculative decoding \citet{leviathan2023fast,chen2023sps} proceeds in three phases. The drafter first samples $B$ candidate tokens $\tilde{x}_t, \ldots, \tilde{x}_{t+B-1}$ from $q_\psi$. The target is then invoked once on the extended sequence and produces, in a single parallel pass, the conditionals $p_\phi(\cdot \mid x_{<t+k})$ for all $k \in \{0, \ldots, B\}$, where $x_{<t+k} \equiv (x_{<t}, \tilde{x}_{t:t+k-1})$. The candidates are finally traversed left-to-right; $\tilde{x}_{t+k}$ is accepted with probability
\begin{equation}
  a_k = \min\!\left(1, \frac{p_\phi(\tilde{x}_{t+k} \mid x_{<t+k})}{q_\psi(\tilde{x}_{t+k} \mid x_{<t+k})}\right),
  \label{eq:accept}
\end{equation}
and on first rejection a replacement is drawn from the residual distribution
\begin{equation}
  x_{t+k} \sim \mathrm{norm}\Big(\max\big(0,\; p_\phi(\cdot \mid x_{<t+k}) - q_\psi(\cdot \mid x_{<t+k})\big)\Big),
  \label{eq:residual}
\end{equation}
terminating the iteration. If all $B$ candidates are accepted, one bonus token is sampled from $p_\phi(\cdot \mid x_{<t+B})$ at no extra cost. The tokens emitted by this procedure are distributed identically to those of standard autoregressive sampling from $p_\phi$ \citet{leviathan2023fast}.

Writing $\alpha_k = \E_{\tilde{x}}[a_k]$ for the marginal acceptance probability at position $k$ and assuming i.i.d.\ rate $\alpha$ across positions, the expected number of tokens emitted per iteration is
\begin{equation}
  \tau(\alpha, B) = \frac{1 - \alpha^{B+1}}{1 - \alpha},
  \label{eq:tau}
\end{equation}
including the bonus token. With $c \in [0,1]$ the wall-clock cost of one drafter pass relative to one target pass, the expected speedup over standard decoding is
\begin{equation}
  S(\alpha, B, c) = \frac{1 - \alpha^{B+1}}{(1 - \alpha)\,(1 + Bc)}.
  \label{eq:speedup}
\end{equation}
$S$ grows with both $B$ and $\alpha$, but each increment to $B$ also incurs additive drafter cost and, typically, a decrease in $\alpha$ at deeper positions \citet{li2024eagle2}. Drafter design reduces to maximising the realised profile $\{\alpha_k\}$ subject to a constraint on $c$.

\subsection{Block-Diffusion Drafters}

Discrete diffusion language models \citet{austin2021d3pm,lou2024sedd,nie2025llada} replace the autoregressive factorization with a denoising formulation: a forward chain progressively masks a clean sequence, and a reverse model trained under bidirectional self-attention emits tokens by iteratively denoising a fully masked block. Block diffusion \citet{arriola2025block} interpolates between this regime and ordinary autoregression by partitioning a sequence into contiguous blocks $b_1, \ldots, b_N$ of size $B$ and factoring
\begin{equation}
  p_\theta(x_{1:T}) = \prod_{n=1}^{N} p_\theta(b_n \mid b_{<n}),
  \label{eq:blockdiff}
\end{equation}
where each block-conditional is parameterised by a discrete diffusion model with bidirectional attention restricted to positions inside $b_n$.

A \emph{block-diffusion drafter} instantiates a block-diffusion model as the speculative drafter $q_\psi$ in the framework of \eqref{eq:accept}--\eqref{eq:speedup}; DFlash \citep{chen2026dflash} is a recent example of this design and the instance we adopt throughout. Following the feature-level drafting principle of EAGLE \citet{li2024eagle,li2025eagle3}, the drafter is conditioned not on discrete tokens but on hidden states extracted from a fixed subset of target layers $\mathcal{L} = \{\ell_1, \ldots, \ell_L\}$. Concretely, given a verified prefix $x_{<t}$ already processed by $p_\phi$, the drafter consumes the multi-layer feature tensor $\Hlt = [h^{(\ell_1)}_{<t} \,\|\, \cdots \,\|\, h^{(\ell_L)}_{<t}]$ and emits the entire $B$-token candidate block in a single non-autoregressive forward pass through a small denoising transformer $f_\psi$ with full self-attention over the $B$ draft positions:
\begin{equation}
  (\tilde{x}_t, \ldots, \tilde{x}_{t+B-1}) \sim q_\psi(\cdot \mid \Hlt) = \mathrm{Cat}\big(f_\psi(\Hlt)\big).
  \label{eq:drafter}
\end{equation}
The parallel block is then fed into the target verification step unchanged. In our experiments we set $B = 16$.

\paragraph{Training objective.} The drafter is trained by teacher-forced position-wise cross-entropy: for a target block $b_n = (x_t, \ldots, x_{t+B-1})$ with corresponding features $\Hlt$,
\begin{equation}
  \LCE(\psi) = -\,\E \sum_{k=1}^{B-1} \log q_\psi(x_{t+k} \mid \Hlt),
  \label{eq:ce}
\end{equation}
where each summand is computed under bidirectional self-attention over the $B$ draft positions. The weighting in \eqref{eq:ce} is uniform across positions, whereas the speedup $S(\alpha, B, c)$ depends on the compounding of left-to-right acceptance probabilities through \eqref{eq:tau}. This mismatch between the symmetric training-time objective and the asymmetric verification-time reward is the central technical problem the remainder of this work addresses.

% ===========================================================================
\begin{figure}[t]
  \centering
  \begin{subfigure}[b]{0.46\linewidth}
    \centering
    \includegraphics[width=\linewidth]{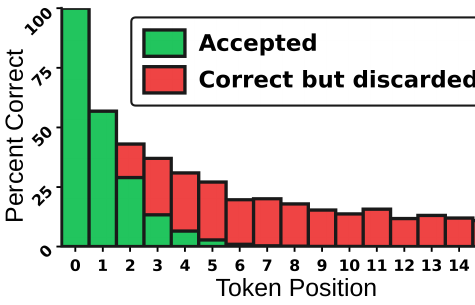}
    \caption{Per-position breakdown on HumanEval.}
    \label{fig:waste-a}
  \end{subfigure}\hfill
  \begin{subfigure}[b]{0.44\linewidth}
    \centering
    \footnotesize
    \begin{tabular}{lccc}
      \toprule
      Benchmark & Accept & Correct & Waste \\
      \midrule
      GSM8K     & 11.6 & 21.9 & 46.9 \\
      MT-Bench  &  9.9 & 17.3 & 42.9 \\
      HumanEval & 13.3 & 27.9 & 52.6 \\
      AIME      & 12.1 & 23.0 & 47.4 \\
      MBPP      & 11.3 & 20.4 & 44.5 \\
      \midrule
      Avg.      & 11.6 & 22.1 & 46.9 \\
      \bottomrule
    \end{tabular}
    \vspace{3.5em}
    \caption{Slot percentages across five benchmarks.}
    \label{fig:waste-b}
  \end{subfigure}
  \caption{Draft-token utilisation of a position-uniform DFlash drafter trained against Llama-3-8B-Instruct with block size $B = 16$. \textbf{(a)} Per-position bars on HumanEval: the green segment is the fraction of draft tokens accepted under the rejection-sampling contract \eqref{eq:accept}; the red segment is the fraction that match the target but are nevertheless discarded as a consequence of an upstream rejection within the same block. Per-position correctness (total bar height) decays only mildly with offset $k$, whereas per-position acceptance (green) decays at the geometric rate implied by \eqref{eq:tau}, so the discarded fraction grows monotonically and dominates beyond a short prefix. \textbf{(b)} The same effect aggregated across benchmarks. Each block contributes 16 candidate slots (15 drafted positions plus the target's bonus correction at the rejection point, which always matches the target's greedy output). \emph{Accept} is the fraction of slots kept by \eqref{eq:accept}; \emph{Correct} is the fraction whose draft (or bonus) matches the target's greedy output; \emph{Waste} is the fraction of those correct tokens that the verification contract discards due to an upstream rejection within the same block, i.e.\ Wasted/(Accepted\,+\,Bonus\,+\,Wasted). All values are percentages.}
  \label{fig:waste}
\end{figure}

% ===========================================================================
\section{The Training--Verification Gap}
\label{sec:gap}
% ===========================================================================

The DFlash training objective \eqref{eq:ce} and the throughput functional \eqref{eq:tau} measure different things. The loss $\LCE$ is a position-wise sum of log-likelihoods under bidirectional self-attention, weighting all $B$ slots equally; the gradient at position $k$ depends on neither the value nor the acceptance of any earlier draft token. The expected accepted length, by contrast, depends on the draft as a whole: position $k$ contributes to $\tau(\alpha, B)$ only when every preceding position is accepted under \eqref{eq:accept}, and that joint event decays geometrically in $k$ whenever per-position acceptance is below one. The training loss therefore spends capacity as if the $B$ positions contributed independently to throughput, while the verifier credits only the longest causally consistent prefix.

Figure~\ref{fig:waste-a} quantifies the resulting waste at the individual-token level. For a standard block-diffusion drafter (DFlash trained against Llama-3-8B-Instruct and evaluated on HumanEval), draft tokens that match the target's greedy output are split by block position into those that verification accepts (green) and those discarded because of an upstream rejection in the same block (red). Per-position correctness decays only mildly with $k$, consistent with the symmetric training signal; per-position acceptance decays at the geometric rate implied by \eqref{eq:tau}, so the red region grows steadily with $k$ and dominates beyond a short prefix.

Figure~\ref{fig:waste-b} confirms that the effect persists across workloads: across GSM8K, MT-Bench, HumanEval, AIME, and MBPP, the drafter's proposals match the target's greedy output in 22.1\% of block slots on average, but only 11.6\% of slots survive the rejection-sampling contract, so an average of 46.9\% of the drafter's correct predictions are discarded as a consequence of upstream rejections, with per-benchmark waste rates between 42.9\% and 52.6\%.

% ===========================================================================
\begin{table}[t]
  \centering
  \small
  \setlength{\tabcolsep}{3.6pt}
  \begin{tabular}{ccccccccccc|c}
    \toprule
    ID & LR & $\gamma$ & $\af$ & $\ac$ & MT-Bench $\tau$ & GSM8K $\tau$ & HumanEval $\tau$ & AIME $\tau$ & MBPP $\tau$ & LiveCode $\tau$ & Avg.\ $\tau\uparrow$ \\
    \midrule
    a & 1e-4 & \None & 0 & 0 & 1.377 & 1.320 & 1.389 & 1.169 & 1.327 & 1.310 & 1.315 \\
    \textbf{b} & \textbf{1e-3} & \textbf{\None} & \textbf{0} & \textbf{0} & \textbf{1.931} & \textbf{2.279} & \textbf{2.803} & \textbf{2.705} & \textbf{2.298} & \textbf{2.241} & \textbf{2.376} \\
    c & 1e-2 & \None & 0 & 0 & 1.670 & 1.905 & 2.319 & 2.031 & 1.907 & 1.861 & 1.949 \\
    \midrule
    d & 1e-3 & 7 & 0 & 0 & 2.050 & 2.289 & 2.949 & 3.017 & 2.239 & 2.232 & 2.463 \\
    \textbf{e} & \textbf{1e-3} & \textbf{10} & \textbf{0} & \textbf{0} & \textbf{2.105} & \textbf{2.382} & \textbf{3.126} & \textbf{3.152} & \textbf{2.412} & \textbf{2.317} & \textbf{2.583} \\
    f & 1e-3 & 20 & 0 & 0 & 2.017 & 2.218 & 2.921 & 2.812 & 2.318 & 2.219 & 2.418 \\
    \midrule
    \textbf{g} & \textbf{1e-3} & \textbf{10} & \textbf{0.3} & \textbf{0} & \textbf{2.222} & \textbf{2.753} & \textbf{3.521} & \textbf{3.465} & \textbf{2.720} & \textbf{2.668} & \textbf{2.892} \\
    h & 1e-3 & 10 & 0.5 & 0 & 2.056 & 2.650 & 3.519 & 3.678 & 2.637 & 2.551 & 2.849 \\
    i & 1e-3 & 10 & 1 & 0 & 2.204 & 2.650 & 3.401 & 3.648 & 2.606 & 2.487 & 2.833 \\
    \midrule
    k & 1e-3 & 10 & 0.3 & 5 & 2.222 & 2.703 & 3.724 & 3.833 & 2.664 & 2.691 & 2.973 \\
    j & 1e-3 & 10 & 0.3 & 10 & 2.185 & 2.762 & 3.957 & 4.278 & 2.904 & 2.769 & 3.143 \\
    l & 1e-3 & 10 & 0.3 & 20 & 2.214 & 2.940 & 3.944 & 4.201 & 2.753 & 2.825 & 3.146 \\
    m & 1e-3 & 10 & 0.3 & 30 & 2.391 & 2.970 & 4.223 & 4.734 & 3.032 & 2.843 & 3.365 \\
    \textbf{n} & \textbf{1e-3} & \textbf{10} & \textbf{0.3} & \textbf{40} & \textbf{2.341} & \textbf{3.074} & \textbf{4.336} & \textbf{4.757} & \textbf{3.088} & \textbf{2.922} & \textbf{3.420} \\
    o & 1e-3 & 10 & 0.3 & 50 & 2.349 & 3.044 & 4.192 & 4.607 & 3.023 & 2.934 & 3.358 \\
    \bottomrule
  \end{tabular}
  \caption{Average acceptance length ($\tau$) across evaluation benchmarks for different training configurations. $\gamma$ is the loss-decay constant; $\af$ is the first-error focal coefficient; and $\ac$ is the chain-loss coefficient. The draft horizon is fixed throughout. Bold entries indicate the best-performing configuration.}
  \label{tab:main}
\end{table}

% ===========================================================================
\section{Training Techniques to Bridge the Gap}
\label{sec:techniques}
% ===========================================================================

The three interventions studied in this section reshape the drafter's training loss along orthogonal axes. Loss decay (Section~\ref{sec:decay}) reweights the per-position contribution to $\LCE$ along the position axis as a hand-specified function of token index. The first-error focal loss (Section~\ref{sec:focal}) is sparser: it adds an auxiliary cross-entropy term restricted to the single position per block that the drafter's argmax decoder currently mispredicts---the \emph{chain breaker} whose rejection truncates the rest of the block---and contributes nothing on blocks the drafter already decodes correctly. The chain reward (Section~\ref{sec:chain}) is the densest of the three along the joint-prefix axis: it augments the loss with a differentiable surrogate for the expected accepted length, so that the gradient at every position is reweighted online by the drafter's current estimate of the prefix-acceptance probability. The interventions are mutually composable; the experiments below layer them in this order.

The peak learning rate is ablated first to factor optimizer dynamics out of the subsequent loss-shape comparisons. Table~\ref{tab:main} reports the average acceptance length $\tau$ on the standard suite for every configuration and is referenced throughout this section; Table~\ref{tab:models} extends the same configurations across Llama-3.2-3B, Llama-3-8B, Qwen-3-4B, and Qwen-3-8B as a cross-target generalisation check.

\subsection{Experimental Setup}
\label{sec:setup}

\paragraph{Target.} All ablations in Table~\ref{tab:main} use Meta-Llama-3-8B-Instruct as the target $p_\phi$: 32 transformer blocks, hidden size 4096, 32 attention heads with 8 KV heads, intermediate size 14{,}336, vocabulary size 128{,}256, RoPE base $5 \times 10^5$ with the Llama-3 long-context scaling.

\paragraph{Drafter.} The drafter $q_\psi$ is a four-layer DFlash \citet{chen2026dflash} denoiser of the architecture introduced in Section~\ref{sec:prelim}: four full-attention transformer blocks, hidden size and head configuration matched to the target ($d = 4096$, 32:8 GQA, $d_{\mathrm{head}} = 128$, $d_{\mathrm{ff}} = 14{,}336$), bfloat16, block size $B = 16$. Multi-layer feature conditioning $\mathcal{L}$ is set to the four evenly-spaced target layers.

\paragraph{Dataset.} Training data is ShareGPT \citep{aeala2023sharegpt}: a compilation of multi-turn user--assistant conversations rendered into Llama-3 chat-template format and truncated at a sequence length of 4{,}096 tokens. Within each sequence, 512 block anchors are sampled per step to construct $(\Hlt, b_n)$ training pairs as defined in \eqref{eq:drafter}. The corpus is not target-distilled: prompts and assistant continuations are taken as-is.

\paragraph{Optimization.} Drafters are trained for 3 epochs with AdamW \citet{loshchilov2019adamw} under a cosine schedule, a warm-up fraction of 1.5\%, and a per-device micro batch of 1 with gradient accumulation of 4, yielding an effective batch size of 32 across the 8 NVIDIA H100 80\,GB GPUs \citep{nvidia2022h100} of a single node. The peak learning rate is the value reported in each row of Table~\ref{tab:main}; unless mentioned, peak LR is fixed at the value selected in Section~\ref{sec:lr}. Training is done using the SpecForge framework \citep{li2025specforge}.

\paragraph{Evaluation.} Acceptance length $\tau$ is averaged over six benchmarks spanning open-ended dialogue \citep[MT-Bench;][]{zheng2023mtbench}, mathematical reasoning (GSM8K, \citealp{cobbe2021gsm8k}; AIME, drawn from the AI-MO validation set of 90 problems from AIME 2022--2024, \citealp{numina2024aime}), and code generation (HumanEval, \citealp{chen2021codex}; MBPP, \citealp{austin2021mbpp}; LiveCodeBench, \citealp{jain2024livecodebench}), reported per-benchmark and as the mean $\taubar$ in the rightmost column of Table~\ref{tab:main}.

\paragraph{Reproducibility.} All experiments are run within a single open-source training framework \citep[SpecForge;][]{li2025specforge} with a single-file drafter definition and per-experiment YAML configurations. Random seeds are fixed per run for anchor sampling, DataLoader shuffling, and dropout; all reported $\taubar$ values are the mean over the last three evaluations of a single training run at the reported step count. Hyperparameters not explicitly ablated (optimizer, warm-up, batch size, effective sequence length, block size, feature-layer set $\mathcal{L}$, drafter architecture) are held fixed across every row of Tables~\ref{tab:main}, \ref{tab:models}, \ref{tab:isolation}, \ref{tab:data}, \ref{tab:blocksize}, \ref{tab:ddtree}, and~\ref{tab:specdiff}, so that any observed $\Delta\taubar$ is attributable to the loss-shape coefficient being ablated in that row.

\subsection{Impact of Learning Rate}
\label{sec:lr}

The peak learning rate of the AdamW optimizer is ablated first with all other hyperparameters held fixed at the configuration of Section~\ref{sec:setup}. Rows a--c of Table~\ref{tab:main} sweep the peak learning rate over $\{10^{-4}, 10^{-3}, 10^{-2}\}$ at the position-uniform baseline ($\gamma = \None$, $\ac = \af = 0$). The optimum sits at $10^{-3}$ (row b), which yields $\taubar = 2.376$ against $1.315$ at $10^{-4}$ (row a) and $1.949$ at $10^{-2}$ (row c)---i.e.\ both an order of magnitude lower and an order of magnitude higher cost the drafter 45\% and 18\% of average acceptance length, respectively. All subsequent experiments fix the peak learning rate at $10^{-3}$.

\subsection{Position-Wise Reweighting: Loss Decay}
\label{sec:decay}

The simplest reconciliation between the position-uniform weight implicit in \eqref{eq:ce} and the geometrically decaying inference value of each position, established in Section~\ref{sec:gap}, is to promote the per-position weight to an explicit hyperparameter. This parameterisation was introduced in the original DFlash work \citep{chen2026dflash}; we adopt it here and study it more systematically as one of three composable interventions. The per-position weight at offset $k \in \{1, \ldots, B-1\}$ is set to

\noindent
\begin{minipage}[t]{0.56\linewidth}
\vspace{0pt}
\begin{equation}
  w_k(\gamma) = \exp\!\left(-\frac{k-1}{\gamma}\right),
  \label{eq:decay}
\end{equation}
parameterised by a decay constant $\gamma > 0$. The reweighted objective then reads
\begin{equation}
  \LCE^{(\gamma)}(\psi) = -\,\E \sum_{k=1}^{B-1} w_k(\gamma) \log q_\psi(x_{t+k} \mid \Hlt).
  \label{eq:ce-decay}
\end{equation}
\end{minipage}\hfill
\begin{minipage}[t]{0.38\linewidth}
\vspace{0pt}
  \centering
  \includegraphics[width=0.95\linewidth]{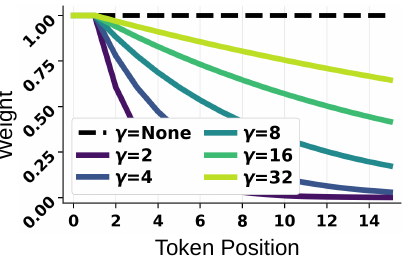}
  \captionof{figure}{Per-position loss weight $w_k(\gamma)$ of \eqref{eq:decay} as a function of the within-block token offset $k \in \{0, \ldots, B-1\}$, plotted for several decay constants $\gamma$ at block size $B = 16$.}
  \label{fig:decay}
\end{minipage}

\medskip
\noindent The limit $\gamma \to \infty$ recovers the position-uniform objective of \eqref{eq:ce}; the limit $\gamma \to 0^+$ collapses to a single-position objective at $k = 1$. Intermediate values induce an exponential profile that, by construction, allocates more gradient signal to the early positions whose acceptance gates the entire suffix and less to the deep positions whose contribution to \eqref{eq:tau} is suppressed.

% ---------------------------------------------------------------------------
\begin{table}[t]
  \centering
  \small
  \setlength{\tabcolsep}{3.4pt}
  \begin{tabular}{lccccccccc|c}
    \toprule
    Model & $\gamma$ & $\af$ & $\ac$ & MT-Bench $\tau$ & GSM8K $\tau$ & HumanEval $\tau$ & AIME $\tau$ & MBPP $\tau$ & LiveCode $\tau$ & Avg.\ $\tau\uparrow$ \\
    \midrule
    \multirow{4}{*}{Llama 3.2-3B}
      & \None & 0 & 0 & 2.351 & 2.875 & 3.778 & 3.003 & 3.110 & 2.546 & 2.944 \\
      & 10 & 0 & 0 & 2.454 & 3.040 & 4.096 & 3.356 & 3.300 & 2.636 & 3.147 \\
      & 10 & 0.3 & 0 & 2.542 & 3.110 & 4.347 & 3.455 & 3.382 & 2.681 & 3.253 \\
      & \textbf{10} & \textbf{0.3} & \textbf{40} & \textbf{2.717} & \textbf{3.507} & \textbf{5.196} & \textbf{4.197} & \textbf{3.933} & \textbf{2.975} & \textbf{3.754} \\
    \midrule
    \multirow{4}{*}{Llama-3-8B}
      & \None & 0 & 0 & 1.931 & 2.279 & 2.803 & 2.705 & 2.298 & 2.241 & 2.376 \\
      & 10 & 0 & 0 & 2.105 & 2.382 & 3.126 & 3.152 & 2.412 & 2.317 & 2.583 \\
      & 10 & 0.3 & 0 & 2.222 & 2.753 & 3.521 & 3.465 & 2.720 & 2.668 & 2.892 \\
      & \textbf{10} & \textbf{0.3} & \textbf{40} & \textbf{2.341} & \textbf{3.074} & \textbf{4.336} & \textbf{4.757} & \textbf{3.088} & \textbf{2.922} & \textbf{3.420} \\
    \midrule
    \multirow{4}{*}{Qwen3-4B}
      & \None & 0 & 0 & 1.875 & 2.397 & 2.738 & 2.133 & 2.396 & 2.265 & 2.301 \\
      & 10 & 0 & 0 & 1.959 & 2.502 & 3.002 & 2.253 & 2.596 & 2.387 & 2.450 \\
      & 10 & 0.3 & 0 & 2.000 & 2.598 & 3.126 & 2.357 & 2.618 & 2.481 & 2.530 \\
      & \textbf{10} & \textbf{0.3} & \textbf{40} & \textbf{2.123} & \textbf{2.833} & \textbf{3.507} & \textbf{2.554} & \textbf{2.958} & \textbf{2.713} & \textbf{2.781} \\
    \midrule
    \multirow{4}{*}{Qwen3-8B}
      & \None & 0 & 0 & 1.836 & 2.248 & 2.480 & 2.030 & 2.032 & 2.144 & 2.128 \\
      & 10 & 0 & 0 & 1.901 & 2.517 & 2.779 & 2.233 & 2.241 & 2.315 & 2.331 \\
      & 10 & 0.3 & 0 & 1.961 & 2.608 & 2.969 & 2.351 & 2.306 & 2.420 & 2.436 \\
      & \textbf{10} & \textbf{0.3} & \textbf{40} & \textbf{2.055} & \textbf{2.795} & \textbf{3.384} & \textbf{2.558} & \textbf{2.466} & \textbf{2.665} & \textbf{2.654} \\
    \bottomrule
  \end{tabular}
  \caption{Acceptance length ($\tau$) across benchmarks for different target-model families and training configurations. $\gamma$ is the loss-decay constant (\None\ denotes the position-uniform objective), $\af$ is the first-error focal coefficient, and $\ac$ is the chain-loss coefficient.}
  \label{tab:models}
\end{table}

This intervention's principal limitation is that the weighting profile \eqref{eq:decay} is fixed and decoupled from the drafter's per-position acceptance behavior at any point in training; every block is downweighted identically, irrespective of its difficulty or of which positions the current drafter is already capable of resolving.

Rows d--f of Table~\ref{tab:main} sweep $\gamma \in \{7, 10, 20\}$ at the tuned learning rate of Section~\ref{sec:lr}; the position-uniform baseline (row b) records $\taubar = 2.376$ as the reference point. Performance peaks at the intermediate setting: $\gamma = 7$ yields $\taubar = 2.463$ (+3.7\%) and $\gamma = 20$ yields $\taubar = 2.418$ (+1.8\%), while $\gamma = 10$ (row e) reaches $\taubar = 2.583$, a +8.7\% gain over the position-uniform baseline.

This matches the reasoning of Section~\ref{sec:gap}: a sharp profile ($\gamma = 7$) cuts off the deep-position signal too aggressively, a slack profile ($\gamma = 20$) is close to the uniform baseline, and the intermediate setting best tracks the decaying inference value of each position implied by \eqref{eq:tau}. We fix $\gamma = 10$ for subsequent ablations.

\subsection{Targeting the Chain Breaker: Focal Loss}
\label{sec:focal}

The decay weighting of Section~\ref{sec:decay} reshapes the gradient with a position-only schedule applied uniformly to every block. The analysis of Section~\ref{sec:gap} suggests a more targeted intervention. Under the rejection-sampling contract \eqref{eq:accept}, a block of $B$ draft tokens contributes to the accepted length up to and including the first position at which the draft is rejected; the identity of that single position---the \emph{chain breaker}---is sufficient to determine the block's acceptance outcome under the argmax simplification. From the perspective of the throughput functional \eqref{eq:tau}, fixing the chain breaker of a block whose prefix is otherwise correct extends its accepted length by exactly one, whereas additional gradient on positions deeper than the breaker yields no improvement until the breaker itself is fixed.

The first-error focal loss instantiates this intuition as an auxiliary cross-entropy term restricted to the chain-breaker position of each block. Concretely, for each block $n$ with draft logits $\ell_{n,k}$ and ground-truth tokens $x^*_{n,k}$, denote the argmax decoder predictions by $\hat{x}_{n,k} = \arg\max_v \ell^v_{n,k}$ and the set of positions where the argmax disagrees with the ground truth by $E_n = \{k \in \{1, \ldots, B-1\} : \hat{x}_{n,k} \neq x^*_{n,k}\}$. Blocks with $E_n = \emptyset$ contribute nothing to the auxiliary term. For blocks with at least one disagreement, let $k^*_n = \min E_n$ be the chain-breaker offset and define
\begin{equation}
  \Lfe(\psi) = \frac{\sum_{n:\, E_n \neq \emptyset} w_{k^*_n}(\gamma)\,\big(-\log q_\psi(x^*_{n,k^*_n} \mid H_{<t_n})\big)}{\sum_{n:\, E_n \neq \emptyset} w_{k^*_n}(\gamma) + \varepsilon},
  \label{eq:focal}
\end{equation}
the mean per-position cross entropy at the chain breaker, weighted by the same position decay $w_k(\gamma)$ used in \eqref{eq:ce-decay}. With $\af \geq 0$ the focal coefficient, the decay-plus-focal objective reads
\begin{equation}
  \mathcal{L}^{(\gamma, \af)}(\psi) = \LCE^{(\gamma)}(\psi) + \af\, \Lfe(\psi).
  \label{eq:decay-focal}
\end{equation}

The construction is qualitatively distinct from the decay weighting. Loss decay applies at every position of every block, redistributing gradient by a smooth function of $k$. The first-error focal term, by contrast, is \emph{block-conditional} (it activates only when the block is breakable) and \emph{position-sparse} (it activates at exactly one offset per such block, selected adaptively by the drafter's current decoder). Two further consequences follow. First, blocks that the drafter already decodes correctly receive no extra signal, which prevents the auxiliary term from over-fitting easy patterns. Second, as the drafter improves, the chain-breaker distribution shifts deeper into the block, so the auxiliary term automatically retargets toward later positions. Rows g--i of Table~\ref{tab:main} sweep $\af \in \{0.3, 0.5, 1.0\}$ on top of the tuned $\gamma = 10$ of Section~\ref{sec:decay} (row e, $\taubar = 2.583$). All three settings improve over the decay-only baseline, and the objective is monotone-decreasing in $\af$ over this range: $\af = 0.3$ (row g) attains $\taubar = 2.892$, a +12.0\% gain over the focal-free decay baseline and a cumulative +21.7\% over the position-uniform row b; $\af = 0.5$ (row h) and $\af = 1.0$ (row i) regress slightly to $\taubar = 2.849$ and $\taubar = 2.833$ respectively. The improvement is largest on the reasoning- and code-heavy benchmarks---HumanEval ($2.803 \to 3.521$, +25.6\%) and AIME ($2.705 \to 3.465$, +28.1\%) relative to row b. We adopt $\af = 0.3$ as the reference setting for the chain sweep that follows.

\subsection{Joint-Prefix Reweighting: Chain Reward}
\label{sec:chain}

The decay weighting of Section~\ref{sec:decay} is a position-only schedule fixed ex ante; the first-error focal loss of Section~\ref{sec:focal} is sparse and block-conditional. A third axis of intervention, complementary to both, is to substitute a differentiable surrogate for the expected accepted length itself, so that the gradient at every position is reweighted online by the prefix-acceptance probability under the current drafter. Let $p_k = q_\psi(x_{t+k} \mid \Hlt)$ denote the drafter's marginal probability on the ground-truth token at offset $k$, and write $\rho_k = \prod_{j=1}^{k} p_j$ for the corresponding prefix probability.

Treating $\rho_k$ as a proxy for the probability that positions $1, \ldots, k$ of the draft are jointly accepted under \eqref{eq:accept}, the expected accepted length is approximated by
\begin{equation}
  \Rchain(\psi) = \frac{1}{B-1} \sum_{k=1}^{B-1} \exp\!\Bigg(\sum_{j=1}^{k} \log p_j\Bigg),
  \label{eq:chain}
\end{equation}
normalised to $[0,1]$. The full training objective combining the three interventions of Sections~\ref{sec:decay}--\ref{sec:chain} is then
\begin{equation}
  \mathcal{L}(\psi) = \LCE^{(\gamma)}(\psi) + \af\, \Lfe(\psi) - \ac\, \Rchain(\psi),
  \label{eq:full}
\end{equation}
with $\ac \geq 0$ controlling the trade-off between marginal correctness and joint chain probability. Differentiating \eqref{eq:chain} with respect to $\log p_k$ yields $(B-1)^{-1} \sum_{j \geq k} \rho_j$: each position is reinforced in proportion to the cumulative prefix probability of its own and every deeper position, so that early positions receive a strictly larger gradient than deep ones, with the relative weighting determined by the drafter's current calibration rather than by a hand-specified schedule.

The implementation reuses the per-token log-probabilities already computed for $\LCE$, since $-\mathrm{CE}(\text{logits}, x^*) = \log p$; the chain reward therefore adds only a cumulative sum and exponentiation per block and does not require a second pass through the language modelling head.

Rows k--o of Table~\ref{tab:main} sweep $\ac \in \{5, 10, 20, 30, 40, 50\}$ on top of the $(\gamma, \af) = (10, 0.3)$ stack of Section~\ref{sec:focal} (row g, $\taubar = 2.892$). Even modest coefficients are markedly accretive: $\ac = 5$ (row k) raises $\taubar$ to 2.973; $\ac = 10$ (row j) to 3.143; $\ac = 20$ (row l) to 3.146; $\ac = 30$ (row m) to 3.365; and $\ac = 40$ (row n) peaks at $\taubar = 3.420$, a +18.3\% gain over the chain-free row g and a cumulative +43.9\% over the position-uniform baseline of row b. The improvement is largest on the reasoning- and code-heavy benchmarks---HumanEval ($2.803 \to 4.336$, +54.7\%) and AIME ($2.705 \to 4.757$, +75.9\%) relative to row b---where long, well-determined draft chains are most plentiful and most rewarded by the joint-prefix reweighting. The objective turns over past the peak: at $\ac = 50$ (row o) $\taubar$ regresses to 3.358, as the chain term begins to dominate the per-position cross entropy. We adopt $\ac = 40$ as the reference setting for the experiments that follow.

\subsection{Experiments on Other Models}
\label{sec:other-models}

Table~\ref{tab:models} replicates the four reference configurations of Section~\ref{sec:setup}---position-uniform baseline ($\gamma = \None$, $\af = \ac = 0$), $+\gamma$, $+\af$, $+\ac$---across four instruction-tuned target models: Llama-3.2-3B-Instruct \citep{metaai2024llama32}, Meta-Llama-3-8B-Instruct \citep{grattafiori2024llama3}, and the instruction-tuned Qwen3-4B and Qwen3-8B releases \citep{yang2025qwen3} ---to confirm that the ranking is target-agnostic. All four are the instruction-tuned variants of their respective families; no separate post-training is performed on the target.

The compounding pattern of Table~\ref{tab:main} reproduces on every target: each successive intervention strictly improves average acceptance length, and the fully stacked $(\gamma, \af, \ac) = (10, 0.3, 40)$ configuration delivers the largest $\taubar$ for all four targets, with cumulative gains over the position-uniform baseline of +27.5\% (Llama-3.2-3B), +43.9\% (Llama-3-8B), +20.9\% (Qwen-3-4B), and +24.7\% (Qwen-3-8B). The absolute gains are largest on the reasoning- and code-heavy benchmarks (HumanEval, AIME) across every target, mirroring the per-benchmark profile of Section~\ref{sec:chain} and indicating that the joint-prefix reweighting transfers target-independently to the workloads on which long, well-determined draft chains are most plentiful.

% ===========================================================================
\section{Ablations}
\label{sec:ablations}
% ===========================================================================

\subsection{Each Technique in Isolation}
\label{sec:isolation}

The sweeps of Section~\ref{sec:techniques} layer the three interventions cumulatively: loss decay is tuned alone, the focal term is added on top of the chosen $\gamma$, and the chain reward is added on top of the decay--plus--focal stack. This construction is the one that yields the strongest configuration of Table~\ref{tab:main}, but it leaves open the question of how much each intervention contributes in isolation---i.e., applied directly to the position-uniform baseline of \eqref{eq:ce} with the other two coefficients held at zero. We isolate each technique by holding the learning rate fixed at the value selected in Section~\ref{sec:lr} and training a separate drafter for each of the techniques.

\begin{wraptable}{r}{0.47\linewidth}
  \vspace{-1em}
  \centering
  \footnotesize
  \setlength{\tabcolsep}{3pt}
  \begin{tabular}{clcc}
    \toprule
    ID & Technique & Setting & Avg.\ $\tau\uparrow$ \\
    \midrule
    a & Position-uniform baseline & --- & 2.376 \\
    b & Loss decay only & $\gamma = 10$ & 2.583 \\
    c & Focal only & $\af = 0.3$ & 2.736 \\
    d & Chain only & $\ac = 40$ & 2.919 \\
    \bottomrule
  \end{tabular}
  \caption{Each training intervention applied in isolation to the position-uniform baseline at the tuned learning rate of Section~\ref{sec:lr}. Row a reproduces row b of Table~\ref{tab:main}; row b reproduces row e; rows c and d are single-technique runs with the other two coefficients held at zero.}
  \label{tab:isolation}
  \vspace{-1em}
\end{wraptable}

Three observations follow from Table~\ref{tab:isolation}. First, every intervention is individually accretive over the position-uniform baseline, and the relative ranking (chain $>$ focal $>$ decay) is consistent with the analysis of Section~\ref{sec:gap}: the more directly an intervention couples its gradient to the verification rule, the larger its isolated effect. Loss decay reweights by a fixed ex-ante schedule that only approximately tracks the geometric decay of acceptance value; the focal term conditions on the drafter's own argmax breaker; the chain reward integrates the joint prefix probability across all positions. Second, the chain reward alone already recovers a large fraction of the fully-stacked gain (row n of Table~\ref{tab:main}, $\taubar = 3.420$), suggesting that the joint-prefix axis is the dominant source of the improvement. Third, the gap between the chain-only configuration and the fully-stacked one nevertheless remains substantial, suggesting that the three interventions do not encode the same training signal.

\subsection{Impact of Training Data}
\label{sec:data}

All experiments of Section~\ref{sec:techniques} use the ShareGPT dataset \citep{aeala2023sharegpt} as the drafter training corpus, taken as-authored rather than re-distilled from the target. To characterize how data composition interacts with the loss-shape interventions of Section~\ref{sec:techniques}, we re-run the four reference configurations on a target-aligned corpus. To provide a diverse mixture of instruction-following and code prompts, following \citet{chen2026dflash}, we collect approximately 800K samples from the NVIDIA Nemotron Post-Training Dataset V2 \citep{nathawani2025nemotron} and CodeAlpaca \citep{chaudhary2023codealpaca}. Rather than training against the as-authored assistant responses, we regenerate each completion under the target model $p_\phi$ (Meta-Llama-3-8B-Instruct, \citealp{grattafiori2024llama3}) served via sglang, yielding a target-aligned training split of the same prompt set. The four reference configurations of Section~\ref{sec:techniques} are then retrained on this split for six epochs at the tuned learning rate and effective batch size of Section~\ref{sec:setup}; metrics are reported at the 75{,}000-step checkpoint.

\begin{wraptable}{r}{0.33\linewidth}
  \vspace{-1em}
  \centering
  \small
  \begin{tabular}{cccc}
    \toprule
    $\gamma$ & $\af$ & $\ac$ & Avg.\ $\tau\uparrow$ \\
    \midrule
    \None & 0 & 0 & 3.914 \\
    10 & 0 & 0 & 4.230 \\
    10 & 0.3 & 0 & 4.633 \\
    \textbf{10} & \textbf{0.3} & \textbf{40} & \textbf{4.985} \\
    \bottomrule
  \end{tabular}
  \caption{Average acceptance length ($\tau$) for the four reference training configurations on the target-aligned Nemotron-V2 + CodeAlpaca split. The best row is bolded.}
  \label{tab:data}
  \vspace{-1em}
\end{wraptable}

Two observations follow from Table~\ref{tab:data}. First, the relative ordering and the cumulative compounding pattern of Section~\ref{sec:techniques} both transfer to the target-aligned data: each successive intervention adds to $\taubar$ (decay +8.1\%, focal a further +9.5\%, chain a further +7.6\%, for a cumulative +27.4\% over the position-uniform baseline), and the fully stacked configuration remains strictly best. Second, the change of training corpus is itself a large independent lever: the position-uniform baseline rises from $\taubar = 2.376$ on ShareGPT (row b of Table~\ref{tab:main}) to $\taubar = 3.914$ on the target-aligned split, a +64.7\% absolute lift before any of the loss-shape interventions are applied; composed with them, the fully stacked configuration reaches $\taubar = 4.985$, a +109.8\% improvement over the ShareGPT position-uniform baseline. The training-time interventions of Section~\ref{sec:techniques} therefore stack multiplicatively with target-aligned training data rather than substituting for it.

\subsection{Impact of Block Size for Evaluation}
\label{sec:blocksize}

\begin{wraptable}{r}{0.56\linewidth}
  \vspace{-1em}
  \centering
  \scriptsize
  \setlength{\tabcolsep}{2pt}
  \begin{tabular}{cccc|ccc}
    \toprule
    \multirow{2}{*}{Block} & \multicolumn{3}{c|}{Base ($\tau\uparrow$)} & \multicolumn{3}{c}{Aligned ($\tau\uparrow$)} \\
    & AIME $\tau$ & HumanEval $\tau$ & MBPP $\tau$ & AIME $\tau$ & HumanEval $\tau$ & MBPP $\tau$ \\
    \midrule
    2  & 1.725 & 1.734 & 1.605 & 1.876 & 1.840 & 1.720 \\
    4  & 2.462 & 2.471 & 2.118 & 3.238 & 3.003 & 2.475 \\
    8  & 2.654 & 2.561 & 2.278 & 4.309 & 3.919 & 2.945 \\
    \textbf{16} & \textbf{2.705} & \textbf{2.803} & \textbf{2.298} & \textbf{4.757} & \textbf{4.336} & \textbf{3.088} \\
    32 & 2.656 & 2.750 & 2.312 & 4.577 & 4.069 & 3.004 \\
    \bottomrule
  \end{tabular}
  \caption{Acceptance length ($\tau$) as a function of the speculation horizon $B$ for the position-uniform baseline (Base) and the fully-aligned $(\gamma, \af, \ac) = (10, 0.3, 40)$ drafter (Aligned). Drafters are trained at $B = 16$ and evaluated at indicated horizon. Best row per column bolded.}
  \label{tab:blocksize}
  \vspace{-1em}
\end{wraptable}

The draft horizon $B$ controls both the per-iteration speedup ceiling \eqref{eq:speedup} and the geometric decay of the per-position acceptance profile $\{\alpha_k\}$. To characterise this trade-off, we evaluate the position-uniform baseline (row b of Table~\ref{tab:main}) and the fully-aligned configuration (row n) at speculation horizons $B \in \{2, 4, 8, 16, 32\}$, holding training at the $B = 16$ block size of Section~\ref{sec:setup} fixed. Table~\ref{tab:blocksize} reports the average acceptance length on AIME, HumanEval, and MBPP.

Two patterns emerge. First, $\taubar$ peaks at the training block size of 16 for both drafters and falls off once the inference horizon exceeds it, with $B = 32$ regressing on every benchmark. The drafter was trained to predict 16-token blocks under bidirectional attention; pushing the inference horizon past that size asks the model to predict positions it never saw during training, and the resulting acceptance loss outweighs the extra speedup ceiling. Second, the aligned drafter's gain over the baseline grows steadily with $B$ below the training horizon: at $B = 2$ the aligned drafter improves AIME $\taubar$ by +8.8\%, at $B = 4$ by +31.5\%, at $B = 8$ by +62.4\%, and at $B = 16$ by +72.0\% (with comparable trajectories on HumanEval and MBPP). The position-uniform baseline stops improving well before $B = 16$ because the deep-position acceptance rate is already near zero; the aligned drafter, whose training signal explicitly targets the early-position acceptance gates and the joint prefix probability, keeps gaining speedup all the way up to the training block size.

\subsection{Application to Inference-time Strategies}
\label{sec:ddtree}

The training-time interventions of Section~\ref{sec:techniques} leave the inference pipeline at the standard rejection-sampling contract: each round verifies a single drafted trajectory of $B$ tokens. A complementary test-time mechanism is to exploit the fact that a single block-diffusion drafter pass already produces a per-position distribution $q_i(\cdot \mid \Hlt)$ at every offset, rather than only the argmax sample. DDTree \citep{ringel2026ddtree} converts these marginals into a draft tree of at most $K$ nodes whose prefix probabilities under the factorised distribution $Q = \prod_i q_i$ are recovered by a best-first heap walk, and verifies the resulting tree in one target-model forward pass with an ancestor-only attention mask. DDTree is wholly orthogonal to the training-time interventions studied here: it changes neither the drafter objective nor the rejection-sampling exactness contract, only the set of continuations the verifier scores per round.

\begin{wraptable}{r}{0.55\linewidth}
  \vspace{-1em}
  \centering
  \scriptsize
  \setlength{\tabcolsep}{3pt}
  \begin{tabular}{lcccc}
    \toprule
    & \multicolumn{2}{c}{No ddTree} & \multicolumn{2}{c}{W/ ddTree} \\
    \cmidrule(lr){2-3}\cmidrule(lr){4-5}
    Technique & Avg.\ $\tau\uparrow$ & +TPS\% $\uparrow$ & Avg.\ $\tau\uparrow$ & +TPS\% $\uparrow$ \\
    \midrule
    Uniform Decay & 2.376 & -- & 3.791 & -- \\
    \quad + Gamma Decay & 2.583 & +12.8\% & 3.999 & +8.6\% \\
    \quad + Focal Loss & 2.892 & +33.3\% & 4.328 & +26.6\% \\
    \quad + Chain Loss & 3.420 & +77.6\% & 4.609 & +56.2\% \\
    \quad + DDTree-Aware Top-$K$ & 3.209 & +66.8\% & 4.751 & +60.1\% \\
    \bottomrule
  \end{tabular}
  \caption{Average accepted-token length ($\tau$) and throughput gain (+TPS\%) relative to the position-uniform baseline. Results are averaged across six benchmarks (GSM8K, HumanEval, AIME25, MBPP, MT-Bench, and LiveCodeBench). The final row trains the DDTree-aware top-$K$ chain reward of Section~\ref{sec:topk} from scratch on the same position-uniform schedule, replacing the greedy chain reward of row n. In this row $K$ denotes the top-$K$ tree width.}
  \label{tab:ddtree}
  \vspace{-1em}
\end{wraptable}

Table~\ref{tab:ddtree} stacks DDTree verification on top of the four reference configurations of Section~\ref{sec:techniques} and reports both acceptance length and throughput gain (relative to the position-uniform baseline) averaged across the six benchmarks. The two axes compose cleanly. Holding the drafter at the position-uniform baseline, DDTree alone raises $\taubar$ from 2.376 to 3.791 (+59.6\%). Holding inference at standard contiguous-block verification, the layered training-time stack raises $\taubar$ from 2.376 to 3.420 (+44.0\%) and throughput by +77.6\% over the position-uniform baseline (last entry of the No-ddTree +TPS\% column). Applied jointly, the fully stacked configuration with DDTree verification reaches $\taubar = 4.609$ at a +56.2\% throughput lift over the position-uniform, no-DDTree baseline (last entry of the W/-ddTree +TPS\% column). Both directions of the cross-product are monotone: every training-time increment retains its gain under DDTree, and DDTree's lift is preserved at every point along the training-time stack. The two surfaces therefore add rather than substitute, consistent with the observation that DDTree reshapes the verifier's search over fixed drafter marginals while the training-time interventions reshape the marginals themselves.

\subsection{DDTree-Aware Top-$K$ Chain Reward}
\label{sec:topk}

The chain reward of Section~\ref{sec:chain} pushes prefix probability mass onto the argmax sequence: each factor of the product $\rho_k = \prod_{j \leq k} p_j$ treats the ground-truth token as the unique survivor at every depth. This is the right training signal when verification follows the single-path rejection-sampling contract of standard speculative decoding, but it under-rewards the broader marginal that DDTree verification \citep{ringel2026ddtree} consumes. Under a tree budget $K$, DDTree expands the heap-best continuation among all prefixes whose factorised probability survives the top-$K$ cut, so a drafter that concentrates all its mass on the single argmax leaves nothing for the tail of plausible continuations DDTree would otherwise verify in parallel.

We introduce a DDTree-aware analogue that rewards the ground-truth token for surviving as one of the top-$K$ candidates per depth (distinct from the block size $B$; here $K$ is the tree width DDTree admits per position), preserving the chain structure of $\Rchain$ while relaxing the argmax-only credit to soft top-$K$ membership. At draft position $k$, let $p_k$ denote the drafter's probability of the ground-truth token and let $p_k^{[K+1]}$ denote the $(K+1)$-th largest probability over the vocabulary at the same position. A soft indicator that the ground-truth token lies in the drafter's top-$K$ at depth $k$ is
\begin{equation}
  s_k^{(K)} = \sigma\!\left(\beta\,\big(p_k - p_k^{[K+1]}\big)\right),
  \label{eq:softtopk}
\end{equation}
with sharpness $\beta > 0$ (we use $\beta = 10$ throughout). The top-$K$ chain reward extends $\Rchain$ from prefix probability under the argmax to prefix probability that the ground truth survives a $K$-wide tree:
\begin{equation}
  \RtopK(\psi) = \frac{1}{B-1} \sum_{k=1}^{B-1} \prod_{j=1}^{k} s_j^{(K)}.
  \label{eq:topk}
\end{equation}
The full DDTree-aware objective adds $\RtopK$ to the chain-reward objective \eqref{eq:full}:
\begin{equation}
  \mathcal{L}_{\mathrm{tK}}(\psi) = \mathcal{L}(\psi) - \atk\, \RtopK(\psi),
  \label{eq:full-topk}
\end{equation}
with $\atk \geq 0$ controlling the trade-off between single-path concentration (favoured by $\Rchain$ at the argmax) and broader top-$K$ coverage (favoured by $\RtopK$ over a width-$K$ tree). The gradient of \eqref{eq:topk} with respect to $\log p_k$ vanishes once $p_k$ comfortably exceeds the $(K+1)$-th quantile (the sigmoid saturates), so capacity is redirected from already-dominant positions toward positions where the ground-truth probability sits near the top-$K$ frontier---exactly the positions whose marginals DDTree's best-first walk discriminates over.

We train from scratch, substituting $\RtopK$ for the greedy chain reward of row n (Table~\ref{tab:main}): the recipe is $(\gamma, \af, \atk, K, \beta) = (10, 0.3, 40, 4, 10)$ with $\ac = 0$, matching the learning rate, batch size, and three-epoch schedule of the chain-reward training of Section~\ref{sec:chain}. The top-$K$ reward redistributes some probability mass away from the argmax toward the next $K-1$ candidates per depth, so single-path acceptance length drops slightly relative to the greedy chain row ($3.420 \to 3.209$, $-6.2\%$ in the No-ddTree column). The mass redistributed onto the top-$K$ continuations is exactly what DDTree exploits, however, and the resulting DDTree-verification $\taubar$ rises from 4.609 to 4.751 (+3.1\% over the already-aligned chain-reward row), more than compensating for the no-tree loss. Per-benchmark, gain is concentrated on reasoning- and code-heavy benchmarks where DDTree's broader marginal matters most: AIME +10.8\% speedup, GSM8K +7.2\%, HumanEval +6.3\%, MBPP +6.1\%, LiveCodeBench +4.8\%; only MT-Bench regresses, by $-2.3\%$, reflecting that benchmark's short, high-entropy responses where the broader marginal carries less reliable signal.

\subsection{Compatibility with SpecDiff-2}
\label{sec:specdiff}

The contemporaneous SpecDiff-2 of \citet{sandler2025specdiff2} targets the same training--verification mismatch we analyse through a different mechanism. Its train-time component, \emph{streak-distillation}, replaces the position-uniform cross-entropy with a pathwise surrogate for the expected accepted streak length: at each prefix $s$, a continuation $x_{1:\gamma}$ is sampled from the frozen verifier $p_\phi$, and the drafter's per-position marginals are scored as $\sum_{m=1}^{\gamma} \prod_{j=1}^{m} q_j(x_j \mid s)$, which is then maximised through the drafter parameters. The outer expectation is taken over teacher prefixes drawn from $p_\phi$, so each gradient step requires a fresh sample of length $\gamma$ from the verifier in addition to the standard drafter pass. The dominant incremental cost of streak-distillation, then, is not the surrogate itself but the verifier rollout that supplies the teacher continuation: training compute scales linearly with the rollout length and, in the published implementation, requires a verifier forward pass per training example above the drafter forward pass already needed for the base loss.

\begin{wraptable}{r}{0.54\linewidth}
  \vspace{-1em}
  \centering
  \scriptsize
  \setlength{\tabcolsep}{1.5pt}
  \begin{tabular}{lccc}
    \toprule
    Technique & LiveCodeBench $\tau\uparrow$ & AIME25 $\tau\uparrow$ & HumanEval $\tau\uparrow$ \\
    \midrule
    Uniform Decay & 2.263 & 2.262 & 2.493 \\
    \quad + Gamma Decay & 2.621 & 2.824 & 2.928 \\
    \quad + Focal Loss & 2.895 & 3.425 & 3.269 \\
    \quad + Chain Loss & 3.361 & 4.650 & 4.202 \\
    \bottomrule
  \end{tabular}
  \caption{Average accepted-token length ($\tau$) on LiveCodeBench, AIME25, and HumanEval for SpecDiff2-trained draft models under progressive training enhancements. Results are reported using the final checkpoint of each training run.}
  \label{tab:specdiff}
  \vspace{-1em}
\end{wraptable}

Our three training-time interventions are compatible with streak-distillation along this axis. The chain reward of Section~\ref{sec:chain} is also a differentiable surrogate for $\E[\text{accepted length}]$, but it is evaluated along the teacher-forced ground-truth tokens already materialised by $\LCE$, so adding it costs only a cumulative sum and exponentiation per block. The decay weighting of Section~\ref{sec:decay} and the focal term of Section~\ref{sec:focal} reuse the same per-token cross-entropy tensor for the same reason. Layering the three interventions on top of streak-distillation therefore inherits the latter's rollout cost but adds nothing further, while reshaping the objective along the position and block-conditional first-error axes that streak-distillation does not address. Table~\ref{tab:specdiff} reports the result of this layering on LiveCodeBench, AIME25, and HumanEval, using SpecDiff-2-trained drafters as the base in every row. Streak-distillation alone (the position-uniform row of the table) records an average $\taubar = 2.339$ on this three-benchmark slice; adding $\gamma = 10$ on top raises $\taubar$ to 2.791, a +19.3\% gain over the streak-distilled baseline; the $\af = 0.3$ focal term adds a further +14.5\% for $\taubar = 3.196$ (+36.6\% cumulative); and the $\ac = 40$ chain reward adds another +27.4\% for a fully stacked $\taubar = 4.071$, a +74.0\% improvement over the streak-distilled baseline.

% ===========================================================================
\section{Conclusion}
% ===========================================================================

We studied the training of block-diffusion drafters for speculative decoding through the lens of the gap between the position-uniform, bidirectional objective inherited from diffusion language modelling and the strictly causal, prefix-truncating acceptance contract that governs verification. A consequence of this gap, visible in the position-uniform DFlash baseline, is that a large majority of the tokens the target ratifies as correct are discarded by the rejection-sampling rule simply because they sit downstream of an earlier within-block rejection---capacity that the standard cross-entropy loss has no mechanism to reclaim.

We analysed three complementary training-time interventions that narrow this gap---position-wise loss decay, a first-error focal loss targeting the chain-breaking position of each block, and a chain reward that substitutes a differentiable surrogate for the expected accepted length. The three reshape the loss along orthogonal axes (position, block-conditional first error, joint prefix), compose additively, preserve the exactness contract of speculative decoding, and add negligible compute over the base objective. They are individually accretive and jointly compounding across the model families and benchmark categories we evaluated, and remain orthogonal to test-time alignment mechanisms developed for the same drafter class. Three concrete follow-ups are suggested by our analysis. First, all three interventions still reshape a teacher-forced loss and do not consume the target distribution at deep positions; combining them with a verifier-side objective (e.g.\ streak-distillation \citep{sandler2025specdiff2} or LK-style acceptance-rate losses \citep{samarin2026lk}) is a natural next step, and our \S\ref{sec:specdiff} composition experiment is a first data point in that direction. Second, the top-$K$ chain reward of \S\ref{sec:topk} suggests that further tailoring the training signal to the specific verifier-side search procedure (tree width, beam schedule) is a promising axis. Third, the transfer of the compounding pattern across four target families and two datasets (Sections~\ref{sec:other-models} and~\ref{sec:data}) argues that a broader systematic study across block-diffusion drafter architectures beyond DFlash should reproduce the same qualitative pattern; we leave that verification to future work.

\bibliography{references}

@misc{aeala2023sharegpt,
  author       = {Aeala and gozfarb and anon8231489123},
  title        = {{ShareGPT\_Vicuna\_unfiltered: A cleaned dump of multi-turn user--ChatGPT conversations}},
  howpublished = {\url{https://huggingface.co/datasets/Aeala/ShareGPT_Vicuna_unfiltered}},
  year         = {2023}
}

@article{agrawal2023sarathi,
  author  = {Amey Agrawal and Ashish Panwar and Jayashree Mohan and Nipun Kwatra and Bhargav S. Gulavani and Ramachandran Ramjee},
  title   = {{SARATHI: Efficient LLM inference by piggybacking decodes with chunked prefills}},
  journal = {arXiv preprint arXiv:2308.16369},
  year    = {2023}
}

@inproceedings{ankner2024hydra,
  author    = {Zachary Ankner and Rishab Parthasarathy and Aniruddha Nrusimha and Christopher Rinard and Jonathan Ragan-Kelley and William Brandon},
  title     = {{Hydra: Sequentially-dependent draft heads for Medusa decoding}},
  booktitle = {Conference on Language Modeling (COLM)},
  year      = {2024}
}

@misc{arriola2025block,
  author = {Marianne Arriola and Aaron Gokaslan and Justin T. Chiu and Zhihan Yang and Zhixuan Qi and Jiaqi Han and Subham Sekhar Sahoo and Volodymyr Kuleshov},
  title  = {{Block diffusion: Interpolating between autoregressive and diffusion language models}},
  year   = {2025}
}

@inproceedings{austin2021d3pm,
  author    = {Jacob Austin and Daniel D. Johnson and Jonathan Ho and Daniel Tarlow and Rianne van den Berg},
  title     = {{Structured denoising diffusion models in discrete state-spaces}},
  booktitle = {Advances in Neural Information Processing Systems (NeurIPS)},
  year      = {2021}
}

@article{austin2021mbpp,
  author  = {Jacob Austin and Augustus Odena and Maxwell Nye and Maarten Bosma and Henryk Michalewski and David Dohan and Ellen Jiang and Carrie Cai and Michael Terry and Quoc Le and Charles Sutton},
  title   = {{Program synthesis with large language models}},
  journal = {arXiv preprint arXiv:2108.07732},
  year    = {2021}
}

@inproceedings{cai2024medusa,
  author    = {Tianle Cai and Yuhong Li and Zhengyang Geng and Hongwu Peng and Jason D. Lee and Deming Chen and Tri Dao},
  title     = {{Medusa: Simple LLM inference acceleration framework with multiple decoding heads}},
  booktitle = {International Conference on Machine Learning (ICML)},
  year      = {2024}
}

@misc{chaudhary2023codealpaca,
  author       = {Sahil Chaudhary},
  title        = {{Code Alpaca: An instruction-following LLaMA model for code generation}},
  howpublished = {\url{https://github.com/sahil280114/codealpaca}},
  year         = {2023}
}

@article{chen2023sps,
  author  = {Charlie Chen and Sebastian Borgeaud and Geoffrey Irving and Jean-Baptiste Lespiau and Laurent Sifre and John Jumper},
  title   = {{Accelerating large language model decoding with speculative sampling}},
  journal = {arXiv preprint arXiv:2302.01318},
  year    = {2023}
}

@inproceedings{chen2026dflash,
  author    = {Jian Chen and Yesheng Liang and Zhijian Liu},
  title     = {{DFlash: Block diffusion for flash speculative decoding}},
  booktitle = {International Conference on Machine Learning (ICML)},
  year      = {2026}
}

@article{chen2021codex,
  author  = {Mark Chen and Jerry Tworek and Heewoo Jun and Qiming Yuan and Henrique Ponde de Oliveira Pinto and Jared Kaplan and Harri Edwards and Yuri Burda and Nicholas Joseph and Greg Brockman and others},
  title   = {{Evaluating large language models trained on code}},
  journal = {arXiv preprint arXiv:2107.03374},
  year    = {2021}
}

@misc{cheng2026dspark,
  author = {Xin Cheng and Xingkai Yu and Chenze Shao and Jiashi Li and Yunfan Xiong and Yi Qian and Jiaqi Zhu and Shirong Ma and Xiaokang Zhang and Jiasheng Ye and Qinyu Chen and Chengqi Deng and Jiping Yu and Damai Dai and Zhengyan Zhang and Yixuan Wei and Yixuan Tan and Wenkai Yang and Runxin Xu and Yu Wu and Zhean Xu and Xuanyu Wang and Muyang Chen and Rui Tian and Xiao Bi and Zhewen Hao and Shaoyuan Chen and Huanqi Cao and Wentao Zhang and Anyi Xu and Huishuai Zhang and Dongyan Zhao and Wenfeng Liang},
  title  = {{Dspark: Confidence-scheduled speculative decoding with semi-autoregressive generation}},
  year   = {2026},
  url    = {https://arxiv.org/abs/2607.05147}
}

@article{cobbe2021gsm8k,
  author  = {Karl Cobbe and Vineet Kosaraju and Mohammad Bavarian and Mark Chen and Heewoo Jun and Lukasz Kaiser and Matthias Plappert and Jerry Tworek and Jacob Hilton and Reiichiro Nakano and Christopher Hesse and John Schulman},
  title   = {{Training verifiers to solve math word problems}},
  journal = {arXiv preprint arXiv:2110.14168},
  year    = {2021}
}

@inproceedings{dao2024flashattention2,
  author    = {Tri Dao},
  title     = {{FlashAttention-2: Faster attention with better parallelism and work partitioning}},
  booktitle = {International Conference on Learning Representations (ICLR)},
  year      = {2024}
}

@inproceedings{dao2022flashattention,
  author    = {Tri Dao and Daniel Y. Fu and Stefano Ermon and Atri Rudra and Christopher R{\'e}},
  title     = {{FlashAttention: Fast and memory-efficient exact attention with IO-awareness}},
  booktitle = {Advances in Neural Information Processing Systems (NeurIPS)},
  year      = {2022}
}

@article{deepseek2025r1,
  author  = {{DeepSeek-AI} and Daya Guo and Dejian Yang and Haowei Zhang and Junxiao Song and Ruoyu Zhang and Runxin Xu and Qihao Zhu and Shirong Ma and Peiyi Wang and others},
  title   = {{DeepSeek-R1: Incentivizing reasoning capability in LLMs via reinforcement learning}},
  journal = {arXiv preprint arXiv:2501.12948},
  year    = {2025}
}

@misc{dettmers2022llmint8,
  author = {Tim Dettmers and Mike Lewis and Younes Belkada and Luke Zettlemoyer},
  title  = {{LLM.int8(): 8-bit matrix multiplication for transformers at scale}},
  year   = {2022}
}

@inproceedings{du2024glide,
  author    = {Cunxiao Du and Jing Jiang and Xu Yuanchen and Jiawei Wu and Sicheng Yu and Yongqi Li and Shengju Li and Kai Xu and Liqiang Nie and Zhaopeng Tu and Yang You},
  title     = {{GliDe with a CaPE: A low-hassle method to accelerate speculative decoding}},
  booktitle = {International Conference on Machine Learning (ICML)},
  year      = {2024}
}

@inproceedings{frantar2023sparsegpt,
  author    = {Elias Frantar and Dan Alistarh},
  title     = {{SparseGPT: Massive language models can be accurately pruned in one-shot}},
  booktitle = {International Conference on Machine Learning (ICML)},
  year      = {2023}
}

@inproceedings{frantar2023gptq,
  author    = {Elias Frantar and Saleh Ashkboos and Torsten Hoefler and Dan Alistarh},
  title     = {{GPTQ: Accurate post-training quantization for generative pre-trained transformers}},
  booktitle = {International Conference on Learning Representations (ICLR)},
  year      = {2023}
}

@inproceedings{fu2024lookahead,
  author    = {Yichao Fu and Peter Bailis and Ion Stoica and Hao Zhang},
  title     = {{Break the sequential dependency of LLM inference using lookahead decoding}},
  booktitle = {International Conference on Machine Learning (ICML)},
  year      = {2024}
}

@techreport{fu2026nemotron,
  author      = {Yonggan Fu and Lexington Whalen and Abhinav Garg and Chengyue Wu and Maksim Khadkevich and Nicolai Oswald and Enze Xie and Daniel Egert and Sharath Turuvekere Sreenivas and Shizhe Diao and Chenhan Yu and Ye Yu and Weijia Chen and Sajad Norouzi and Jingyu Liu and Shiyi Lan and Ligeng Zhu and Jin Wang and Jindong Jiang and Morteza Mardani and Mehran Maghoumi and Song Han and Ante Juki{\'c} and Nima Tajbakhsh and Jan Kautz and Pavlo Molchanov},
  title       = {{Nemotron-labs-diffusion: A tri-mode language model unifying autoregressive, diffusion, and self-speculation decoding}},
  institution = {NVIDIA},
  year        = {2026},
  note        = {Technical report}
}

@misc{fu2026efficientdlm,
  author = {Yonggan Fu and Lexington Whalen and Zhifan Ye and Xin Dong and Shizhe Diao and Jingyu Liu and Chengyue Wu and Hao Zhang and Enze Xie and Song Han and Maksim Khadkevich and Jan Kautz and Yingyan Celine Lin and Pavlo Molchanov},
  title  = {{Efficient-dlm: From autoregressive to diffusion language models, and beyond in speed}},
  year   = {2026},
  url    = {https://arxiv.org/abs/2512.14067}
}

@inproceedings{ghazvininejad2019maskpredict,
  author    = {Marjan Ghazvininejad and Omer Levy and Yinhan Liu and Luke Zettlemoyer},
  title     = {{Mask-Predict: Parallel decoding of conditional masked language models}},
  booktitle = {Proceedings of the 2019 Conference on Empirical Methods in Natural Language Processing (EMNLP)},
  year      = {2019}
}

@inproceedings{ghazvininejad2020axe,
  author    = {Marjan Ghazvininejad and Vladimir Karpukhin and Luke Zettlemoyer and Omer Levy},
  title     = {{Aligned cross entropy for non-autoregressive machine translation}},
  booktitle = {International Conference on Machine Learning (ICML)},
  year      = {2020}
}

@misc{ghazvininejad2020smart,
  author = {Marjan Ghazvininejad and Omer Levy and Luke Zettlemoyer},
  title  = {{Semi-autoregressive training improves mask-predict decoding}},
  year   = {2020}
}

@misc{gong2023diffuseq,
  author = {Shansan Gong and Mukai Li and Jiangtao Feng and Zhiyong Wu and Lingpeng Kong},
  title  = {{DiffuSeq: Sequence to sequence text generation with diffusion models}},
  year   = {2023}
}

@article{grattafiori2024llama3,
  author  = {Aaron Grattafiori and Abhimanyu Dubey and Abhinav Jauhri and Abhinav Pandey and Abhishek Kadian and Ahmad Al-Dahle and Aiesha Letman and Akhil Mathur and Alan Schelten and Alex Vaughan and others},
  title   = {{The Llama 3 herd of models}},
  journal = {arXiv preprint arXiv:2407.21783},
  year    = {2024}
}

@inproceedings{gu2018nat,
  author    = {Jiatao Gu and James Bradbury and Caiming Xiong and Victor O. K. Li and Richard Socher},
  title     = {{Non-autoregressive neural machine translation}},
  booktitle = {International Conference on Learning Representations (ICLR)},
  year      = {2018}
}

@inproceedings{gulrajani2023plaid,
  author    = {Ishaan Gulrajani and Tatsunori B. Hashimoto},
  title     = {{Likelihood-based diffusion language models}},
  booktitle = {Advances in Neural Information Processing Systems (NeurIPS)},
  year      = {2023}
}

@article{hinton2015distilling,
  author  = {Geoffrey Hinton and Oriol Vinyals and Jeff Dean},
  title   = {{Distilling the knowledge in a neural network}},
  journal = {arXiv preprint arXiv:1503.02531},
  year    = {2015}
}

@misc{huang2026domino,
  author = {Jianuo Huang and Yaojie Zhang and Qituan Zhang and Hao Lin and Hanlin Xu and Linfeng Zhang},
  title  = {{Domino: Decoupling causal modeling from autoregressive drafting in speculative decoding}},
  year   = {2026},
  url    = {https://arxiv.org/abs/2605.29707}
}

@misc{incoai2026dflash2,
  author = {{Inco AI}},
  title  = {{DFlash 2: Keep Drafting Parallel}},
  month  = aug,
  year   = {2026},
  url    = {https://inco.ai/blog/dflash2/}
}

@article{jain2024livecodebench,
  author  = {Naman Jain and King Han and Alex Gu and Wen-Ding Li and Fanjia Yan and Tianjun Zhang and Sida Wang and Armando Solar-Lezama and Koushik Sen and Ion Stoica},
  title   = {{LiveCodeBench: Holistic and contamination free evaluation of large language models for code}},
  journal = {arXiv preprint arXiv:2403.07974},
  year    = {2024}
}

@inproceedings{kwon2023vllm,
  author    = {Woosuk Kwon and Zhuohan Li and Siyuan Zhuang and Ying Sheng and Lianmin Zheng and Cody Hao Yu and Joseph E. Gonzalez and Hao Zhang and Ion Stoica},
  title     = {{Efficient memory management for large language model serving with PagedAttention}},
  booktitle = {Proceedings of the 29th Symposium on Operating Systems Principles (SOSP)},
  year      = {2023}
}

@inproceedings{lee2018iterative,
  author    = {Jason Lee and Elman Mansimov and Kyunghyun Cho},
  title     = {{Deterministic non-autoregressive neural sequence modeling by iterative refinement}},
  booktitle = {Proceedings of the 2018 Conference on Empirical Methods in Natural Language Processing (EMNLP)},
  year      = {2018}
}

@inproceedings{leviathan2023fast,
  author    = {Yaniv Leviathan and Matan Kalman and Yossi Matias},
  title     = {{Fast inference from transformers via speculative decoding}},
  booktitle = {International Conference on Machine Learning (ICML)},
  year      = {2023}
}

@misc{li2025specforge,
  author       = {Shenggui Li and Yikai Zhu and Chao Wang and Fan Yin and Shuai Shi and Yubo Wang and Yi Zhang and Yingyi Huang and Haoshuai Zheng and Yineng Zhang},
  title        = {{SpecForge: Train speculative decoding models effortlessly and port them smoothly to SGLang serving}},
  howpublished = {\url{https://github.com/sgl-project/SpecForge}},
  year         = {2025}
}

@inproceedings{li2022diffusionlm,
  author    = {Xiang Lisa Li and John Thickstun and Ishaan Gulrajani and Percy Liang and Tatsunori B. Hashimoto},
  title     = {{Diffusion-LM improves controllable text generation}},
  booktitle = {Advances in Neural Information Processing Systems (NeurIPS)},
  year      = {2022}
}

@inproceedings{li2024eagle,
  author    = {Yuhui Li and Fangyun Wei and Chao Zhang and Hongyang Zhang},
  title     = {{EAGLE: Speculative sampling requires rethinking feature uncertainty}},
  booktitle = {International Conference on Machine Learning (ICML)},
  year      = {2024}
}

@inproceedings{li2024eagle2,
  author    = {Yuhui Li and Fangyun Wei and Chao Zhang and Hongyang Zhang},
  title     = {{EAGLE-2: Faster inference of language models with dynamic draft trees}},
  booktitle = {Proceedings of the 2024 Conference on Empirical Methods in Natural Language Processing (EMNLP)},
  year      = {2024}
}

@article{li2025eagle3,
  author  = {Yuhui Li and Fangyun Wei and Chao Zhang and Hongyang Zhang},
  title   = {{EAGLE-3: Scaling up inference acceleration of large language models via training-time test}},
  journal = {arXiv preprint arXiv:2503.01840},
  year    = {2025}
}

@inproceedings{lin2024awq,
  author    = {Ji Lin and Jiaming Tang and Haotian Tang and Shang Yang and Wei-Ming Chen and Wei-Chen Wang and Guangxuan Xiao and Xingyu Dang and Chuang Gan and Song Han},
  title     = {{AWQ: Activation-aware weight quantization for LLM compression and acceleration}},
  booktitle = {Proceedings of Machine Learning and Systems (MLSys)},
  year      = {2024}
}

@misc{liu2024kangaroo,
  author = {Fangcheng Liu and Yehui Tang and Zhenhua Liu and Yunsheng Ni and Kai Han and Yunhe Wang},
  title  = {{Kangaroo: Lossless self-speculative decoding via double early exiting}},
  year   = {2024}
}

@misc{liu2025tidar,
  author = {Jingyu Liu and Xin Dong and Zhifan Ye and Rishabh Mehta and Yonggan Fu and Vartika Singh and Jan Kautz and Ce Zhang and Pavlo Molchanov},
  title  = {{Tidar: Think in diffusion, talk in autoregression}},
  year   = {2025},
  url    = {https://arxiv.org/abs/2511.08923}
}

@inproceedings{liu2024online,
  author    = {Xiaoxuan Liu and Lanxiang Hu and Peter Bailis and Ion Stoica and Zhijie Deng and Alvin Cheung and Hao Zhang},
  title     = {{Online speculative decoding}},
  booktitle = {International Conference on Machine Learning (ICML)},
  year      = {2024}
}

@misc{loshchilov2019adamw,
  author = {Ilya Loshchilov and Frank Hutter},
  title  = {{Decoupled weight decay regularization}},
  year   = {2019},
  url    = {https://arxiv.org/abs/1711.05101}
}

@inproceedings{lou2024sedd,
  author    = {Aaron Lou and Chenlin Meng and Stefano Ermon},
  title     = {{Discrete diffusion modeling by estimating the ratios of the data distribution}},
  booktitle = {International Conference on Machine Learning (ICML)},
  year      = {2024}
}

@misc{metaai2024llama32,
  author       = {{Meta AI}},
  title        = {{Llama 3.2: Revolutionizing edge AI and vision with open, customizable models}},
  howpublished = {\url{https://ai.meta.com/blog/llama-3-2-connect-2024-vision-edge-mobile-devices/}},
  year         = {2024}
}

@inproceedings{miao2024specinfer,
  author    = {Xupeng Miao and Gabriele Oliaro and Zhihao Zhang and Xinhao Cheng and Zeyu Wang and Zhengxin Zhang and Rae Ying Yee Wong and Alan Zhu and Lijie Yang and Xiaoxiang Shi and Chunan Shi and Zhuoming Chen and Daiyaan Arfeen and Reyna Abhyankar and Zhihao Jia},
  title     = {{SpecInfer: Accelerating large language model serving with tree-based speculative inference and verification}},
  booktitle = {Proceedings of the 29th ACM International Conference on Architectural Support for Programming Languages and Operating Systems (ASPLOS)},
  year      = {2024}
}

@misc{nathawani2025nemotron,
  author       = {Devvrit Nathawani and {NVIDIA}},
  title        = {{Nemotron post-training dataset V2}},
  howpublished = {\url{https://huggingface.co/datasets/nvidia/Nemotron-Post-Training-Dataset-v2}},
  year         = {2025}
}

@article{nie2025llada,
  author  = {Shen Nie and Fengqi Zhu and Zebin You and Xiaolu Zhang and Jingyang Ou and Jun Hu and Jun Zhou and Yankai Lin and Ji-Rong Wen and Chongxuan Li},
  title   = {{Large language diffusion models}},
  journal = {arXiv preprint arXiv:2502.09992},
  year    = {2025}
}

@misc{nvidia2022h100,
  author       = {{NVIDIA Corporation}},
  title        = {{NVIDIA H100 tensor core GPU architecture}},
  howpublished = {\url{https://resources.nvidia.com/en-us-tensor-core/gtc22-whitepaper-hopper}},
  year         = {2022}
}

@article{openai2024o1,
  author  = {{OpenAI}},
  title   = {{OpenAI o1 system card}},
  journal = {arXiv preprint arXiv:2412.16720},
  year    = {2024}
}

@misc{paszke2019pytorch,
  author = {Adam Paszke and Sam Gross and Francisco Massa and Adam Lerer and James Bradbury and Gregory Chanan and Trevor Killeen and Zeming Lin and Natalia Gimelshein and Luca Antiga and Alban Desmaison and Andreas K{\"o}pf and Edward Yang and Zach DeVito and Martin Raison and Alykhan Tejani and Sasank Chilamkurthy and Benoit Steiner and Lu Fang and Junjie Bai and Soumith Chintala},
  title  = {{Pytorch: An imperative style, high-performance deep learning library}},
  year   = {2019},
  url    = {https://arxiv.org/abs/1912.01703}
}

@misc{pope2023efficiently,
  author = {Reiner Pope and Sholto Douglas and Aakanksha Chowdhery and Jacob Devlin and James Bradbury and Jonathan Heek and Kefan Xiao and Shivani Agrawal and Jeff Dean},
  title  = {{Efficiently scaling transformer inference}},
  year   = {2023}
}

@misc{numina2024aime,
  author       = {{Project Numina (AI-MO)}},
  title        = {{AIMO validation AIME: 90 american invitational mathematics examination problems from AIME 2022--2024}},
  howpublished = {\url{https://huggingface.co/datasets/AI-MO/aimo-validation-aime}},
  year         = {2024}
}

@inproceedings{qian2021glat,
  author    = {Lihua Qian and Hao Zhou and Yu Bao and Mingxuan Wang and Lin Qiu and Weinan Zhang and Yong Yu and Lei Li},
  title     = {{Glancing transformer for non-autoregressive neural machine translation}},
  booktitle = {Proceedings of the 59th Annual Meeting of the Association for Computational Linguistics (ACL)},
  year      = {2021}
}

@article{ringel2026ddtree,
  author  = {Liran Ringel and Yaniv Romano},
  title   = {{Accelerating speculative decoding with block diffusion draft trees}},
  journal = {arXiv preprint arXiv:2604.12989},
  year    = {2026}
}

@article{roziere2023codellama,
  author  = {Baptiste Rozi{\`e}re and Jonas Gehring and Fabian Gloeckle and Sten Sootla and Itai Gat and Xiaoqing Ellen Tan and Yossi Adi and Jingyu Liu and Tal Remez and J{\'e}r{\'e}my Rapin and others},
  title   = {{Code Llama: Open foundation models for code}},
  journal = {arXiv preprint arXiv:2308.12950},
  year    = {2023}
}

@inproceedings{samarin2026lk,
  author    = {Alexander Samarin and Sergei Krutikov and Anton Shevtsov and Sergei Skvortsov and Filipp Fisin and Alexander Golubev},
  title     = {{LK losses: Direct acceptance rate optimization for speculative decoding}},
  booktitle = {International Conference on Machine Learning (ICML)},
  year      = {2026}
}

@article{sandler2025specdiff2,
  author  = {Jameson Sandler and Jacob K. Christopher and Thomas Hartvigsen and Ferdinando Fioretto},
  title   = {{SpecDiff-2: Scaling diffusion drafter alignment for faster speculative decoding}},
  journal = {arXiv preprint arXiv:2511.00606},
  year    = {2025}
}

@article{sanh2019distilbert,
  author  = {Victor Sanh and Lysandre Debut and Julien Chaumond and Thomas Wolf},
  title   = {{DistilBERT, a distilled version of BERT: smaller, faster, cheaper and lighter}},
  journal = {arXiv preprint arXiv:1910.01108},
  year    = {2019}
}

@inproceedings{savinov2022sundae,
  author    = {Nikolay Savinov and Junyoung Chung and Mikolaj Binkowski and Erich Elsen and A{\"a}ron van den Oord},
  title     = {{Step-unrolled denoising autoencoders for text generation}},
  booktitle = {International Conference on Learning Representations (ICLR)},
  year      = {2022}
}

@inproceedings{stern2018blockwise,
  author    = {Mitchell Stern and Noam Shazeer and Jakob Uszkoreit},
  title     = {{Blockwise parallel decoding for deep autoregressive models}},
  booktitle = {Advances in Neural Information Processing Systems (NeurIPS)},
  year      = {2018}
}

@inproceedings{sun2024wanda,
  author    = {Mingjie Sun and Zhuang Liu and Anna Bair and J. Zico Kolter},
  title     = {{A simple and effective pruning approach for large language models}},
  booktitle = {International Conference on Learning Representations (ICLR)},
  year      = {2024}
}

@inproceedings{wolf2020transformers,
  author    = {Thomas Wolf and Lysandre Debut and Victor Sanh and Julien Chaumond and Clement Delangue and Anthony Moi and Perric Cistac and Clara Ma and Yacine Jernite and Julien Plu and Canwen Xu and Teven Le Scao and Sylvain Gugger and Mariama Drame and Quentin Lhoest and Alexander M. Rush},
  title     = {{Transformers: State-of-the-art natural language processing}},
  booktitle = {Proceedings of the 2020 Conference on Empirical Methods in Natural Language Processing: System Demonstrations},
  pages     = {38--45},
  address   = {Online},
  publisher = {Association for Computational Linguistics},
  year      = {2020},
  url       = {https://aclanthology.org/2020.emnlp-demos.6/}
}

@article{wu2026dpace,
  author  = {Tianyu Wu and Yu Yao and Zhenting Qi and Han Zheng and Zhuohan Wang and Haoran Ma and Lawrence Liao and Himabindu Lakkaraju and Ju Li and Yilun Du},
  title   = {{D-PACE: Dynamic position-aware cross-entropy for parallel speculative drafting}},
  journal = {arXiv preprint arXiv:2605.18810},
  year    = {2026}
}

@inproceedings{xiao2023smoothquant,
  author    = {Guangxuan Xiao and Ji Lin and Mickael Seznec and Hao Wu and Julien Demouth and Song Han},
  title     = {{SmoothQuant: Accurate and efficient post-training quantization for large language models}},
  booktitle = {International Conference on Machine Learning (ICML)},
  year      = {2023}
}

@article{yang2025qwen3,
  author  = {An Yang and Anfeng Li and Baosong Yang and Beichen Zhang and Binyuan Hui and Bo Zheng and Bowen Yu and Chengyuan Gao and Chengen Huang and Chenxuan Lv and others},
  title   = {{Qwen3 technical report}},
  journal = {arXiv preprint arXiv:2505.09388},
  year    = {2025}
}

@inproceedings{yu2022orca,
  author    = {Gyeong-In Yu and Joo Seong Jeong and Geon-Woo Kim and Soojeong Kim and Byung-Gon Chun},
  title     = {{Orca: A distributed serving system for transformer-based generative models}},
  booktitle = {16th USENIX Symposium on Operating Systems Design and Implementation (OSDI)},
  year      = {2022}
}

@inproceedings{zhang2024draftverify,
  author    = {Jun Zhang and Jue Wang and Huan Li and Lidan Shou and Ke Chen and Gang Chen and Sharad Mehrotra},
  title     = {{Draft \& verify: Lossless large language model acceleration via self-speculative decoding}},
  booktitle = {Proceedings of the 62nd Annual Meeting of the Association for Computational Linguistics (ACL)},
  year      = {2024}
}

@article{zhang2024hass,
  author  = {Lefan Zhang and Xiaodan Wang and Yanhua Huang and Ruiwen Xu},
  title   = {{Learning harmonized representations for speculative sampling}},
  journal = {arXiv preprint arXiv:2408.15766},
  year    = {2024}
}

@inproceedings{zheng2023mtbench,
  author    = {Lianmin Zheng and Wei-Lin Chiang and Ying Sheng and Siyuan Zhuang and Zhanghao Wu and Yonghao Zhuang and Zi Lin and Zhuohan Li and Dacheng Li and Eric P. Xing and Hao Zhang and Joseph E. Gonzalez and Ion Stoica},
  title     = {{Judging LLM-as-a-judge with MT-Bench and Chatbot Arena}},
  booktitle = {Advances in Neural Information Processing Systems (NeurIPS), Datasets and Benchmarks Track},
  year      = {2023}
}

@inproceedings{zhou2024distillspec,
  author    = {Yongchao Zhou and Kaifeng Lyu and Ankit Singh Rawat and Aditya Krishna Menon and Afshin Rostamizadeh and Sanjiv Kumar and Jean-Fran{\c{c}}ois Kagy and Rishabh Agarwal},
  title     = {{DistillSpec: Improving speculative decoding via knowledge distillation}},
  booktitle = {International Conference on Learning Representations (ICLR)},
  year      = {2024}
}
\bibliographystyle{tmlr}

% ===========================================================================
\clearpage
\section{Appendix}
\label{sec:appendix}
% ===========================================================================

\begin{table}[!htb]
  \caption{We show that our techniques are complementary to D-PACE. At matched smoothing, adding focal+chain improves D-PACE's own configuration on both target models, by +3.1\% on Llama-3-8B-Instruct and +2.9\% on Qwen3-8B-Instruct, in each case yielding the best result for that target. $\Delta\taubar$ is relative to D-PACE at matched smoothing.}
  \label{tab:dpace}
  \centering
  \small
  \setlength{\tabcolsep}{4.5pt}
  \begin{tabular}{llcccccccc}
    \toprule
    Target & Method & MTB & GSM8K & HEval & AIME & MBPP & LCB & $\taubar$ & $\Delta\taubar$ \\
    \midrule
    \multirow{2}{*}{Llama-3-8B}
      & D-PACE (matched reference) & 2.199 & 3.227 & 4.336 & 4.889 & 3.178 & 2.984 & 3.469 & --- \\
      & \quad + focal + chain ($\gamma$ off) & 2.197 & 3.246 & 4.543 & 5.231 & 3.215 & 3.033 & 3.578 & +3.1\% \\
    \midrule
    \multirow{2}{*}{Qwen3-8B}
      & D-PACE (matched reference) & 1.825 & 2.907 & 3.293 & 2.919 & 2.549 & 2.484 & 2.663 & --- \\
      & \quad + focal + chain ($\gamma$ off) & 1.790 & 2.976 & 3.455 & 3.068 & 2.615 & 2.540 & 2.741 & +2.9\% \\
    \bottomrule
  \end{tabular}
\end{table}

\subsection{Difference between our chain-reward objective and D-PACE}
\label{app:dpace}

Here we clarify the differences between the chain-reward objective introduced in Section~\ref{sec:chain} and the objective outlined in D-PACE \citep{wu2026dpace}. Both objectives derive from the same prefix-product surrogate of expected accepted length, $S(\psi) = \sum_k \prod_{i \leq k} q_i$, and both reduce to the same per-position coefficient $\sum_{m \geq j} \prod_{i \leq m} q_i$. D-PACE applies this coefficient ($\alpha_s$) as a detached importance weight multiplying the per-position cross-entropy. Because the weight is a scalar, this reproduces $\nabla_\psi S$ when the smoothing $\alpha_s = 0$: which is why their $\alpha_s = 0$ row and their surrogate-only Accept-rate baseline both collapse to $\tau \approx 1.02$ (their Tables 1 and 5).

The main difference between our technique and theirs is the placement. Under multiplicative reweighting, a poorly calibrated drafter drives $\prod_{i \leq j} q_i \to 0$ and the entire loss vanishes at every position simultaneously; D-PACE floors this with asymmetric smoothing $\tilde{q}_i = (1 - \alpha_s) q_i + \alpha_s$ (Section 3.3 of D-PACE). Furthermore, the choice of $\alpha_s$ for D-PACE is very important, as the $\alpha_s = 0$ case above illustrates: without a nonzero floor, detachment collapses and training reduces to the degenerate surrogate-only case. We instead compose additively: $-\ac \Rchain$ is placed on top of an unscaled per-token cross-entropy (optionally with our position decay and first-error focal terms), so the CE supplies a per-token acceptance anchor that cannot be scaled away and no smoothing floor is required.

Two consequences follow from the choice of placement. First, $\alpha_s$ and $\ac$ play different roles: $\alpha_s$ is a numerical-stability floor that must be nonzero for training to proceed at all, whereas $\ac$ is a trade-off coefficient with an interior optimum (Table~\ref{tab:main}). Second, because $\Rchain$ is normalised by $(B-1)$, $\ac$ is decoupled from block size, whereas D-PACE's weights $\bar{w}_j = \sum_{m=j}^{B} \prod_{i \leq m} \tilde{q}_i$ are unnormalized, coupling the effective magnitude of the base loss to block size.

Furthermore, our techniques are stackable on top of D-PACE. Table~\ref{tab:dpace} shows that adding our focal and chain terms to D-PACE at matched smoothing $\alpha_s = 0.5$ improves on D-PACE alone ($\taubar = 3.469 \to 3.578$).

\subsection{Proposed techniques under TV/LK style losses}
\label{app:tvlk}

The three interventions of Sections~\ref{sec:decay}--\ref{sec:chain} are defined in terms of the drafter's per-position log-probabilities on the target-selected tokens, not in terms of the cross-entropy itself. They can therefore compose with any per-token loss family that exposes those quantities. We verify this by replacing the CE loss with two acceptance-oriented alternatives: a total-variation (TV) objective and the LK loss of \citet{samarin2026lk}. We then re-run the full stack on top of each and record acceptance lengths for both Llama3-8B-Instruct and Qwen3-8B-Instruct. Table~\ref{tab:tvlk} reports the comparison for both targets. We see that in all cases, our proposed techniques improve acceptance lengths. On Llama-3-8B-Instruct the stack adds +48.3\% over TV and +51.1\% over LK; on Qwen3-8B-Instruct, +16.7\% and +21.7\% respectively. With a refit of the chain-reward coefficient $\ac = 10$, the TV-based stack further improves to 3.568 on Llama-3-8B-Instruct. Thus, our techniques are not limited to CE losses, and consistently show improvement on all losses tested.

\begin{table}[!htb]
  \caption{The proposed stack is not specific to cross-entropy. Average emitted length $\taubar$ per benchmark for three per-token base losses, each with and without the full stack (position decay + first-error focal + chain reward), on both targets. $\Delta\taubar$ is relative to the corresponding base loss.}
  \label{tab:tvlk}
  \centering
  \small
  \begin{tabular}{lcccccccc}
    \toprule
    Method & MTB & GSM8K & HEval & AIME & MBPP & LCB & $\taubar$ & $\Delta\taubar$ \\
    \midrule
    \multicolumn{9}{l}{\emph{Llama-3-8B-Instruct}} \\
    TV only & 1.793 & 1.924 & 2.653 & 2.610 & 2.180 & 2.024 & 2.197 & --- \\
    \quad + stack & 2.082 & 2.960 & 4.180 & 4.553 & 2.922 & 2.853 & 3.258 & +48.3\% \\
    \quad + stack ($\ac$ refit 10) & 2.282 & 3.270 & 4.545 & 5.030 & 3.247 & 3.033 & 3.568 & +62.4\% \\
    LK only & 1.696 & 2.116 & 2.674 & 2.569 & 2.122 & 2.186 & 2.227 & --- \\
    \quad + stack & 2.079 & 3.088 & 4.223 & 4.826 & 3.018 & 2.954 & 3.365 & +51.1\% \\
    \midrule
    \multicolumn{9}{l}{\emph{Qwen3-8B-Instruct}} \\
    TV only & 1.691 & 2.575 & 2.828 & 2.453 & 2.319 & 2.277 & 2.357 & --- \\
    \quad + stack & 1.839 & 2.960 & 3.463 & 3.051 & 2.622 & 2.564 & 2.750 & +16.7\% \\
    LK only & 1.651 & 2.438 & 2.615 & 2.300 & 2.207 & 2.153 & 2.227 & --- \\
    \quad + stack & 1.813 & 2.932 & 3.390 & 2.985 & 2.604 & 2.542 & 2.711 & +21.7\% \\
    \bottomrule
  \end{tabular}
\end{table}

\subsection{Stability of Acceptance Length Over Training}
\label{app:stability}

\begin{figure}[!htb]
  \centering
  \includegraphics[width=0.9\linewidth]{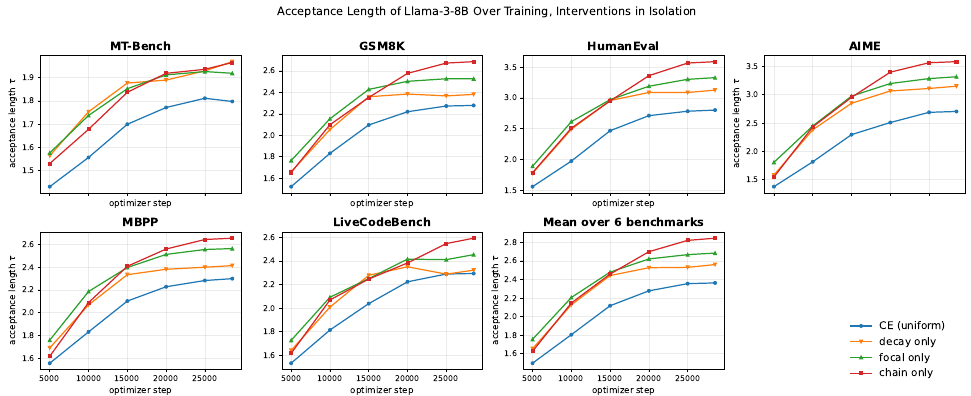}
  \caption{Acceptance lengths using Llama-3-8B-Instruct as our target model, each technique in isolation.}
  \label{fig:llama-iso}
\end{figure}

\begin{figure}[!htb]
  \centering
  \includegraphics[width=0.9\linewidth]{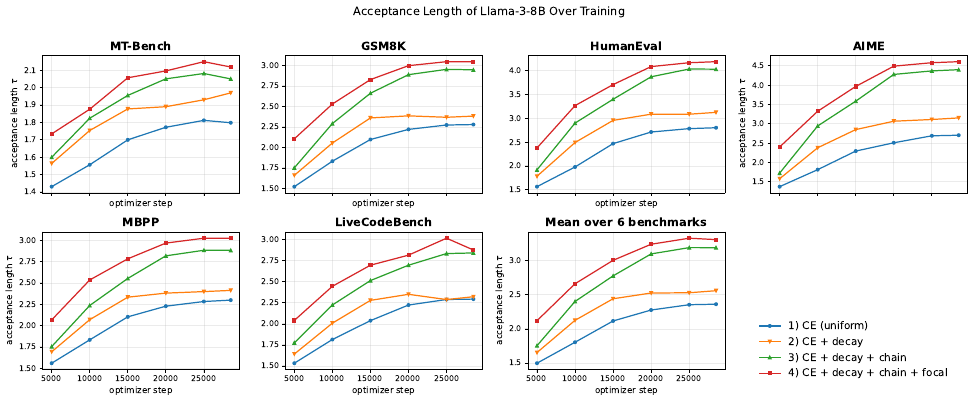}
  \caption{Acceptance lengths using Llama-3-8B-Instruct as our target model, progressively stacking each technique.}
  \label{fig:llama-stack}
\end{figure}

\begin{figure}[!htb]
  \centering
  \includegraphics[width=0.9\linewidth]{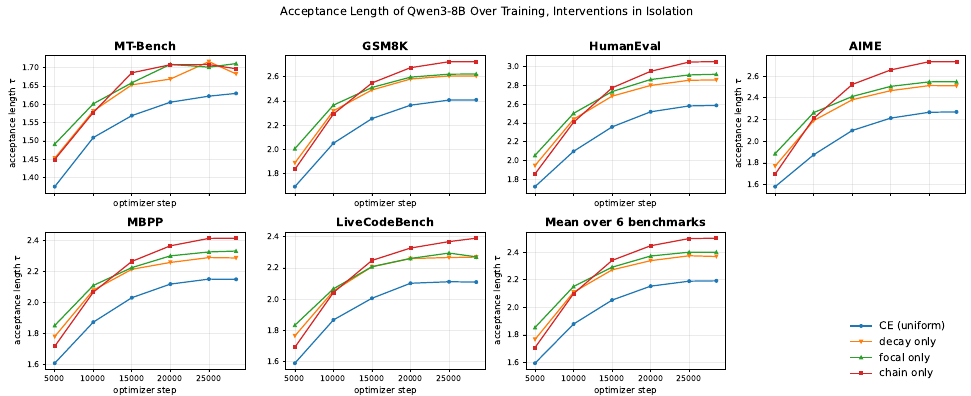}
  \caption{Acceptance lengths using Qwen-3-8B-Instruct as our target model, each technique in isolation.}
  \label{fig:qwen-iso}
\end{figure}

\begin{figure}[!htb]
  \centering
  \includegraphics[width=0.9\linewidth]{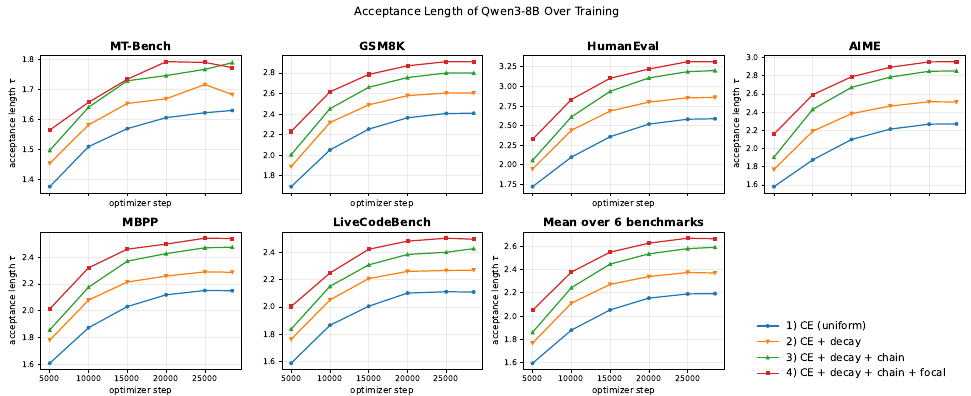}
  \caption{Acceptance lengths using Qwen-3-8B-Instruct as our target model, progressively stacking each technique.}
  \label{fig:qwen-stack}
\end{figure}

In the following figures \ref{fig:llama-iso}, \ref{fig:llama-stack}, \ref{fig:qwen-iso}, and \ref{fig:qwen-stack}, we show how acceptance length is impacted over training. When considering each intervention in isolation, we see that all offer improvements. For the case of Llama3-8B-Instruct, the decay intervention in isolation offers an 8.4\% improvement over the baseline ($2.363 \to 2.561$), the focal intervention offers 13.6\% (2.685), and the chain reward offers the largest individual gain of 20.4\% (2.845). When incrementally stacking the interventions, the gains are cumulative and consistent at every checkpoint: adding decay to uniform CE improves acceptance length by 8.4\%, adding the chain reward on top brings a further 24.7\% ($2.561 \to 3.192$), and adding the focal term completes the stack at 3.310 --- a total of +40.1\% over the CE baseline and +16.3\% above the best single intervention, with the full gap already present early in training (+41.6\% at 5k steps).

For the case of Qwen3-8B-Instruct, the same picture holds: in isolation, decay, focal, and chain improve over the 2.192 baseline by 8.1\%, 9.5\%, and 14.1\% (2.369, 2.401, 2.502), while the incremental stack reaches 2.369 (+decay), 2.591 (+decay+chain), and 2.664 (full stack), for +21.5\% over baseline and +6.5\% over the best isolated arm. We see that for both model families, stacking in general offers better stability of acceptance lengths over training, and larger improvements than each technique in isolation.

\subsection{Positioning relative to prior non-autoregressive generation works}
\label{app:nar}

Here we elaborate on our work's positioning relative to non-autoregressive (NAR) generation works. Mask-Predict \citep{ghazvininejad2019maskpredict} drops the decoder's causal attention mask and decodes in a constant number of iterations, re-masking the lowest-confidence tokens and re-predicting them from the surviving context. SMART \citep{ghazvininejad2020smart} builds training inputs from the model's own predictions rather than ground-truth tokens, so training sees the potentially erroneous context inference supplies. AXE \citep{ghazvininejad2020axe} replaces position-wise cross entropy with a dynamic program over the best monotonic alignment. GLAT \citep{qian2021glat} reveals target tokens adaptively, at a rate set by the model's current error rate.

All four close a gap in output quality: parallel generation degrades accuracy relative to autoregressive decoding, and these methods recover part of it.

Under our speculative decoding setting, verification of a drafted token is credited only if the entire prefix ahead of it has been accepted, so per-token accuracy is necessary but not sufficient (i.e, tokens cannot be treated independently, see Section~\ref{sec:gap}).

We further clarify why the interventions proposed in these works cannot be transferred to our setting. First, there is no repair pass. Mask-Predict (and SMART, which uses it at decoding time) can correct a wrong token at a later iteration, so the model conditions on its own earlier mistakes, which is what SMART's prediction-based context addresses; GLAT decodes in a single pass, where a wrong token degrades the output without affecting its neighboring tokens. In both cases the remedy assumes that an error is either recoverable or local. Neither holds under verification as the target scores the block once, truncates at the first rejection, and never revisits the accepted prefix, so an error will impact the suffix.

Our three interventions instead leave conditioning and targets untouched and reshape how much each position's loss counts, each responding to a different consequence of truncation (i.e, focus on the dependency of tokens in the verification setting). Position decay (Section~\ref{sec:decay}) weights early positions more heavily because their acceptance gates every token behind them. The first-error focal term (Section~\ref{sec:focal}) focuses on the first incorrect predicted token, as this will cause all tokens after it (regardless of correctness) to be thrown away. The chain reward (Section~\ref{sec:chain}) optimizes a differentiable surrogate for prefix survival rather than a sum of independent per-position losses. Thus, our setting focuses on verification and dependency among tokens, while other works focus on individual token correctness without the verification constraint.

\subsection{Comparison with existing speculative decoding systems}
\label{app:eagle}

Our paper primarily regards optimizations for diffusion/block-based drafters. However, it is insightful to also compare against EAGLE-2 \citep{li2024eagle2} and EAGLE-3 \citep{li2025eagle3}, strong autoregressive drafters. We train both drafters from scratch under the same training conditions described in Section~\ref{sec:setup}, and evaluate all methods under vLLM \citep{kwon2023vllm} harness for obtaining TPS. Note that in prior experiments, we used standard Pytorch \citep{paszke2019pytorch}/transformers \citep{wolf2020transformers}-based evaluation, whereas vLLM has optimizations specific to decoding. EAGLE runs use the official dynamic-tree configuration (depth 5, top-k 10) at two candidate budgets: 60, its default, and 16, which matches the drafting budget of our $B = 16$ block. On Llama-3-8B-Instruct, our stacked drafter accepts more tokens per step than EAGLE-2 at either budget ($\taubar = 3.32$ versus 2.85 at budget 60 and 2.51 at 16), and does so while verifying 16 candidates rather than 60.

\begin{table}[!htb]
  \centering
  \small
  \begin{tabular}{lcccccccc}
    \toprule
    Method & MTB & GSM8K & HEval & AIME & MBPP & LCB & $\taubar$ & $\times$AR \\
    \midrule
    AR baseline & 158.8 & 156.8 & 162.2 & 161.9 & 150.9 & 158.3 & 1.00 & 1.00$\times$ \\
    EAGLE-2$_{60}$ & 204.6 & 234.7 & 256.9 & 206.1 & 240.0 & 229.4 & 2.85 & 1.45$\times$ \\
    EAGLE-2$_{16}$ & 203.3 & 219.7 & 238.3 & 191.1 & 223.9 & 218.1 & 2.51 & 1.37$\times$ \\
    EAGLE-3$_{60}$ & 220.5 & 258.6 & 247.6 & 256.1 & 216.7 & 201.8 & 2.54 & 1.48$\times$ \\
    EAGLE-3$_{16}$ & 219.5 & 239.7 & 246.2 & 239.8 & 205.8 & 199.3 & 2.27 & 1.42$\times$ \\
    DFlash (CE) & 214.6 & 291.2 & 346.8 & 342.6 & 282.0 & 294.8 & 2.36 & 1.87$\times$ \\
    DFlash (stack) & 245.2 & 378.8 & 520.0 & 589.4 & 364.2 & 364.2 & 3.32 & 2.59$\times$ \\
    \bottomrule
  \end{tabular}
  \caption{Throughput (tok/s) for different speculative decoding methods using Llama-3-8B-Instruct as target measured under vLLM. Speedups over the AR baseline are reported in the last column; per-benchmark speedups follow from the baseline row. EAGLE drafters are trained in-house under the conditions of Section~\ref{sec:setup}; subscripts denote draft-tree budget.}
  \label{tab:throughput}
\end{table}

The stacked drafter reaches 2.59$\times$ AR against EAGLE-2$_{60}$'s 1.45$\times$ and EAGLE-2$_{16}$'s 1.37$\times$. EAGLE-3 trained under the same conditions behaves similarly (1.48$\times$ at budget 60, 1.42$\times$ at 16). In all cases, our proposed techniques yield both higher acceptance rates and better speedups.

\begin{figure}[!htb]
  \centering
  \includegraphics[width=0.9\linewidth]{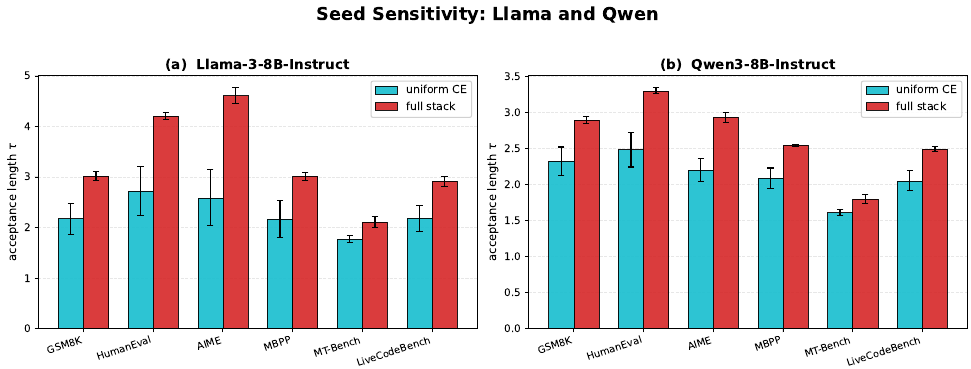}
  \caption{Seed sensitivity of the position-uniform baseline and the fully stacked objective on Llama-3-8B-Instruct and Qwen3-8B-Instruct. Each point is the mean acceptance length over three training seeds $\{42, 2024, 7\}$; whiskers are 95\% confidence intervals.}
  \label{fig:seeds}
\end{figure}

\subsection{Multiple Run Confidence Intervals}
\label{app:seeds}

To ensure that the improvements gained from the techniques are not an artifact of training variance, we retrained the two main configurations, the position-uniform baseline and the fully stacked objective $(\gamma, \af, \ac) = (10, 0.3, 40)$, under three random seeds $\{42, 2024, 7\}$ on both Llama-3-8B-Instruct and Qwen3-8B-Instruct, under the same training settings as Section~\ref{sec:setup}. Each run is evaluated at its final checkpoint on the full six-benchmark suite. Figure~\ref{fig:seeds} reports per-benchmark means with 95\% confidence intervals. On Llama-3-8B the stack raises mean acceptance length from 2.268 to 3.313, a gap of +1.045 against a combined seed standard deviation of 0.138; on Qwen3-8B the gap is +0.534 against 0.067. The confidence intervals for $\taubar$ are disjoint on both targets by a wide margin ([1.943, 2.593] versus [3.295, 3.331] on Llama-3-8B, [1.973, 2.273] versus [2.638, 2.675] on Qwen3-8B), and the separation holds per benchmark: across all twelve target $\times$ benchmark pairs, the two configurations are separated by more than one standard deviation, the closest being MT-Bench on Qwen3-8B. The experiments show that the interventions' improvements hold over a variety of different seeds.

\subsection{Accepted Length versus End-to-End Throughput}
\label{app:throughput}

Here we discuss the difference between acceptance length and end-to-end throughput. As this paper primarily is concerned with optimizations under the same architecture / training settings, we mainly report acceptance lengths as these are more transferrable across different hardware environments. However, it is important to note that higher acceptance lengths do not always lead to better end-to-end throughput.

The conversion from accepted length to wall-clock throughput depends on drafter cost per step, verification cost, speculation horizon, and draft structure. Table~\ref{tab:throughput} illustrates such decoupling. DFlash (CE) accepts fewer tokens per step than EAGLE-2$_{60}$ ($\taubar = 2.36$ versus 2.85) yet is substantially faster end-to-end (1.87$\times$ versus 1.45$\times$ AR), primarily due to the fact that we can draft a full block of tokens when using diffusion/block-based drafters.

We therefore report throughput as the primary metric in Table~\ref{tab:throughput}, and treat $\taubar$ throughout the paper as the diagnostic that explains it: accepted length for the controlled training comparisons, tokens per second for every claim about deployment.

\begin{table}[!htb]
  \centering
  \small
  \begin{tabular}{llccc}
    \toprule
    Architecture & Configuration & $\taubar$ & $\Delta\taubar$ & Rel. \\
    \midrule
    DFlash-2 & CE-uniform & 2.079 & --- & --- \\
    DFlash-2 & + full stack & 2.883 & +0.804 & +38.7\% \\
    Domino & default & 3.240 & --- & --- \\
    Domino & + full stack & 3.897 & +0.657 & +20.3\% \\
    DSpark (semi-AR) & stock recipe & 2.459 & --- & --- \\
    DSpark (semi-AR) & + full stack & 2.816 & +0.357 & +14.5\% \\
    \bottomrule
  \end{tabular}
  \caption{Composition of our training objectives with drafter architectures that model causal structure directly. Target model Llama-3-8B-Instruct.}
  \label{tab:arch}
\end{table}

\subsection{Composition with Architecture-Side Causal Correction}
\label{app:arch}

A complementary line of work supplies causal structure through the drafter architecture rather than the training objective. DFlash-2\citep{incoai2026dflash2} injects a dynamic short convolution over draft features, Domino \citep{huang2026domino} attaches a lightweight GRU head that carries an explicit causal path through the block, and DSpark \citep{cheng2026dspark} schedules semi-autoregressive decoding by confidence. Since our objectives target the same intra-block blindness from the training side, whether the two compose or substitute is a natural question. We measured this on three drafter architectures under the evaluation settings of Section~\ref{sec:setup}.

On each architecture, we see benefits from our techniques. DFlash-2 gains +38.7\%, Domino gains +20.3\%, DSpark, gains +14.5\%. We note that while all models were trained under the same settings of Section~\ref{sec:setup}, due to varying architectures, it is possible that there are better settings than the ones tested. We leave such search for future work.

\end{document}